\documentclass{article}

\usepackage{arxiv}

\usepackage[utf8]{inputenc} %
\usepackage[T1]{fontenc}    %
\usepackage{hyperref}       %
\usepackage{url}            %
\usepackage{booktabs}       %
\usepackage{amsfonts}       %
\usepackage{nicefrac}       %
\usepackage{microtype}      %
\usepackage{graphicx}
\usepackage{comment}
\usepackage{tabularx}
\usepackage{chngcntr}
\usepackage{physics}
\usepackage{multirow}
\usepackage{braket}
\usepackage{tablefootnote}
\usepackage{cleveref}
\usepackage{amsmath}
\usepackage[giveninits=true, backend=biber, style=numeric-comp, sorting=none, minbibnames=5, maxbibnames=7, isbn=false,url=true,eprint=false]{biblatex}
\bibliography{references.bib}

\usepackage{doi}
\usepackage{xcolor}
\usepackage[acronym]{glossaries}
\usepackage{longtable}
\usepackage{float}

\title{Machine Learning Benchmarks for the Classification of Equivalent Circuit Models from Electrochemical Impedance Spectra}

    \author{
    Joachim Schaeffer \\
	CCPS Laboratory \\
	TU Darmstadt, Germany \\
	\texttt{\scriptsize joachim.schaeffer@tu-darmstadt.de} \\
	\And
	Paul Gasper \\
	National Renewable Energy Lab\\
	Golden, CO, USA\\
	\texttt{\scriptsize paul.gasper@nrel.gov}
 	\And
	Esteban Garcia-Tamayo\\
	Titan Advanced Energy Solutions\\
	Salem, MA, USA\\
	\texttt{\scriptsize e.garciatamayo@gmail.com} 
 	\And
	Raymond Gasper \\
	Kingston, MA, USA\\
	\texttt{\scriptsize raymondgasper@fastmail.com}
	\And
	Masaki Adachi \\
	Machine Learning Research Group\\
	University of Oxford, UK\\
	\texttt{\scriptsize masaki@robots.ox.ac.uk}
	\And
	Juan Pablo Gaviria-Cardona \\
	Universidad Pontificia Bolivariana \\
	Medellin, Colombia\\
	\texttt{\scriptsize juanpablo.gaviria@upb.edu.co} 
 	\And
	Simon Montoya-Bedoya \\
	Verasonics SAS \\
	Medellin, Colombia\\
	\texttt{\scriptsize simonmontoya@verasonics.com} 
 	\And
	Anoushka Bhutani\\
	Department of Mechanical Engineering \\
	Carnegie Mellon University\\
	Pittsburgh, PA, USA\\
	\texttt{\scriptsize anoushkb@andrew.cmu.edu} 
 	\And
	Andrew Schiek \\
	National Renewable Energy Lab\\
	Golden, CO, USA\\
	\texttt{\scriptsize andrew.schiek@nrel.gov}
	\And
	Rhys Goodall \\
	Chemix.ai \\
	Sunnyvale, CA, USA \\
	\texttt{{\scriptsize rhys.goodall@chemix.ai}}
	\And
	\hspace{1.3cm} Rolf Findeisen \\
	\hspace{1.3cm} CCPS Laboratory \\
	\hspace{1.3cm} TU Darmstadt, Germany \\
	\hspace{1.3cm} \texttt{\scriptsize rolf.findeisen@tu-darmstadt.de} \\
	\And
	Richard D. Braatz \\ 
	Massachusetts Institute of Technology\\
	Cambridge, MA, USA\\
	\texttt{\scriptsize braatz@mit.edu} 
	\And
	Simon Engelke \\
	Battery Associates \\
	Dublin, Ireland\\
	\texttt{\scriptsize simon.engelke@battery.associates}
}

\renewcommand{\headeright}{}
\renewcommand{\undertitle}{}
\renewcommand{\shorttitle}{Machine Learning Benchmarks for Classification of Electrochemical Impedance Spectra}

\hypersetup{
pdftitle={ML Electrochemical Impedance Spectroscopy},
pdfsubject={},
pdfauthor={},
pdfkeywords={Electrochemical Impedance Spectroscopy, Machine Learning, Equivalent Circuit Model, Classification, Lithium-Ion Batteries, Hackathon, Open Data},
}
\newacronym{cnn}{CNN}{Convolutional Neural Network}
\newacronym{ecm}{ECM}{Equivalent Circuit Model}
\newacronym{eis}{EIS}{Electrochemical Impedance Spectroscopy}
\newacronym{qs}{QS}{QuantumScape}
\newacronym{ml}{ML}{Machine Learning}
\newacronym{umap}{UMAP}{Uniform Manifold Approximation and Projection}
\newacronym{rf}{RF}{Random Forest}
\newacronym{lda}{LDA}{Linear Discriminant Analysis}
\newacronym{qda}{QDA}{Quadratic Discriminant Analysis}
\newacronym{lib}{LIBs}{lithium-ion batteries}
\newacronym{soh}{SOH}{State-Of-Health}
\newacronym{soc}{SOC}{State-Of-Charge}
\newacronym{xgboost}{XGBoost}{extreme gradient boosting}

\immediate\write18{cp `kpsewhich unsrt.bst` .}
\begin{document}
\maketitle
\begin{abstract}
    Analysis of \gls{eis} data for electrochemical systems often consists of defining an \gls{ecm} using expert knowledge and then optimizing the model parameters to deconvolute various resistance, capacitive, inductive, or diffusion responses. For small data sets, this procedure can be conducted manually; however, it is not feasible to manually define a proper \gls{ecm} for extensive data sets with a wide range of \gls{eis} responses. Automatic identification of an \gls{ecm} would substantially accelerate the analysis of large sets of \gls{eis} data. We showcase machine learning methods to classify the \glspl{ecm} of 9,300 impedance spectra provided by QuantumScape for the BatteryDEV hackathon. The best-performing approach is a gradient-boosted tree model utilizing a library to automatically generate features, followed by a random forest model using the raw spectral data. A convolutional neural network using boolean images of Nyquist representations is presented as an alternative, although it achieves a lower accuracy.
    We publish the data and open source the associated code. The approaches described in this article can serve as benchmarks for further studies. A key remaining challenge is the identifiability of the labels, underlined by the model performances and the comparison of misclassified spectra.
\end{abstract} 
\glsreset{eis} 
\glsreset{ecm} 

\keywords{Electrochemical Impedance Spectroscopy \and Machine Learning \and Equivalent Circuit Model \and Classification \and Lithium-Ion Batteries \and Hackathon \and Open Data}

\section{Introduction}
\label{sec:intro}
Processes inside \gls{lib} and other electrochemical devices occur at different timescales \cite{krewer2018dynamic}. In \gls{lib}, lithium ions are shuttled between positive and negative electrodes, via the electrolyte and separator, mostly through diffusion processes. The kinetics during this transition vary due to differences between lithium-ion diffusion coefficients in liquids (electrolyte) and solids (positive/negative electrode active materials), which give rise to the different timescales mentioned. One method for monitoring the various responses of electrochemical systems over different timescales is \gls{eis}, a non-invasive technique that uses AC voltage or current signals over a spectrum of frequencies to excite processes within the electrochemical system. These spectra can thus facilitate the evaluation of electrochemical systems \cite{Wang2021EISRev}, such as batteries \cite{choi2020modeling,andre2011characterization,batmodel_eis2016,krewer2018dynamic}, fuel cells \cite{niya2013eisfuelcells}, supercapacitors \cite{supercapeis2015}, corrosion \cite{bonora1996eiscorrosion}, or biological systems \cite{randviir2013eisbioanalyt}. For batteries, particular research areas exploiting \gls{eis} are \gls{ecm} characterization \cite{choi2020modeling}, blocking electrode experiments to investigate purely ionic or electronic behaviors \cite{qian2001eisblockingelec}, diffusion processes modeling \cite{oldenburger2019investigation}, characterization of porous electrodes \cite{ogihara2012theoretical}, electrode characterization via transmission line modeling \cite{abarbanel2015exploring}, and monitoring of cell performance \cite{ZhangLionDegEIS2020}. However, to analyze these spectra and to assign a specific mechanism such as electronic resistance, charge-transfer, mass transport, etc., electrochemists usually employ \gls{ecm} to represent the different physicochemical processes in the battery by parameterizing them in terms of electrical circuit elements such as inductance, resistance, capacitance, or a combination of them. Defining the structure of an \gls{ecm} generally requires expert judgment, meaning that evaluation of a very large number of \gls{eis} measurements is a difficult process to automate.

As in many other scientific and engineering fields, \Gls{ml} methods have become popular in the area of electrochemistry to accelerate data analysis or modeling tasks, especially for large data sets. For example, \gls{ml} methods have been used successfully for predicting the remaining useful life of batteries both in laboratory environments \cite{severson2019data} as well as in deployed systems \cite{aitioSolar2021}. Furthermore, machine learning methods gained popularity during the last years for analyzing spectral data such as FTIR spectra \cite{kedzierski2019ftir_ml_microplas, lavadeschafferbraatz2022}, Raman spectra \cite{lavadeschafferbraatz2022, ralbovsky2020raman_ml, lussier2020deep_raman}, X-ray diffraction spectra \cite{suzuki2020symmetry}, and \gls{eis} data \cite{zhu2019equivalent, interpreatable_ml_eis2022, puthongkham2021machine, bongiorno2022ml_corrosion, xu2020eis_ecoli_xgb, ZhangLionDegEIS2020, jones_penelope_2021_5704796}. Recent developments include Bayesian model selection for \gls{eis} data \cite{adachi2022bayesian, adachi2023sober} based on fast Bayesian inference using quadrature \cite{adachi2022fast}.
\cite{https://doi.org/10.48550/arxiv.1704.04861}
Further development of software tools and \gls{ml} methods for analyzing impedance data can be accelerated by the publication of open-source software libraries and data sets. There exists open-source software to analyze \gls{eis} data (e.g. \cite{Murbach_2018_EIS_tool, bayesDRT2021}). However, there is still significant potential and need for analysis software and machine learning approaches to be shared and published open source. The situation is similar for data. Various data sets are available \cite{ZhangLionDegEIS2020, kollmeyer_2022, mohtat2021reversible, jones_penelope_2021_5704796}, but the total amount of open data is still small compared to the wide range of applications and the diversity of \gls{eis} data that arises from those. So, this article aims to contribute to the growing body of open-source battery data and software. We focus on using \gls{ml} methods to accurately classify the latent \gls{ecm}.  
With this article, we share a large synthetic \gls{eis} data set with the associates \glspl{ecm} and an unlabeled data set consisting of synthetic and measured data, both provided by \gls{qs} for the international BatteryDEV hackathon to attract a variety of researchers and source interdisciplinary solutions for the problem of \gls{ecm} identification.

\paragraph{BatteryDEV Hackathon} The machine learning approaches described in this article were partially developed during the one-week open-source BatteryDEV hackathon in March 2022. However, additional approaches were developed afterward, and the existing work was refined for this article. This publication makes the data and the code publicly available. The BatteryDEV hackathons were started by its host organization Battery Associates to foster innovation in the battery space. The first BatteryDEV hackathon took place in January 2021, and the second iteration in March 2022. The BatteryDEV hackathons receive support from industry and academia, as described in the acknowledgments. Within the context of batteries, the objectives of BatteryDEV are to (1) increase global collaboration involving data across sectors, (2) encourage the development of open-source solutions for analyzing data, and (3) provide an opportunity for hands-on training to grow the pool of global talent. 
For BatteryDEV 2022, there were 140 registrations, 85 people joined the hackathon, and there were submissions from 60 participants. There were registrations from more than 20 countries, and participants included Data Analytics, \gls{ml}, Battery, and Energy Materials experts from industry and academia, many participating in interdisciplinary teams. The need for more openly available data sets in the battery space is widely accepted \cite{attia2020bayesian,ward2021principles}. Hackathons can accelerate innovation and have been shown to yield exciting results in other fields \cite{crowd_sourcing_materials}. 

This article is organized as follows. Section \ref{sec:data} describes the \gls{eis} data set provided by \gls{qs}. Section \ref{sec:challenge_description} defines the purpose and the challenges of the hackathon. Section \ref{sec:approaches} reports the approaches followed by Section \ref{sec:discussion} for discussion of other ideas and Section \ref{sec:conclusion} outlines the paper conclusions. 

\section{EIS Data Set}
\label{sec:data}
The \gls{eis} data set, $\mathcal{Z}$, associated with this article was created by \gls{qs}. It comprises approximately 9,300 synthetic impedance spectra, $\mathbf{z} \in \mathcal{Z}$ that are vectors of impedances, $\mathbf{z} \in \mathbb{C}^{n}$. Furthermore, the data set includes the generating \gls{ecm} types. No noise was added to the data. The parameters of the \glspl{ecm} were drawn from independent reciprocal distributions, except for the time constants, which were drawn from uniform distributions. Furthermore, it was ensured that the time constants of the RC elements are significantly different for each impedance spectrum associated with the RC-RC-RCPE-RCPE circuit, i.e., $\max(\{\tau_1/\tau_2, \tau_2/\tau_1\}) > 10$. The parameter bounds and frequency ranges are informed by the solid-state battery R\&D of \gls{qs}. Furthermore, \gls{qs} shared an unlabeled data set containing 80\% synthetic and 20\% measured data, which is also made available but not analyzed in this article. Information about how to access the data can be found in the data sharing section.

Table ~\ref{ECM} illustrates the nine classes of the predefined \glspl{ecm}. A hyphen denotes series connections, and a combination of two elements denotes a parallel connection. Each parallel connection has only two branches, each consisting of one element. The name in parenthesis is a shorthand notation. The number of impedance spectra in parenthesis is the number of spectra prior to filtering out spectra that met filter criteria described in the Supplementary Information Sec.\,\ref{sec-si-dapre-filter}. These criteria remove spectra that are either unphysical or unlikely to represent a physical battery or both. However, for completeness, we include results for the filtered and unfiltered data in Tab.\,\ref{tab:class_res_overview}. All other results correspond to the filtered data. 
Table ~\ref{element} states the six circuit elements, names, number of parameters, and equations for each of the elements which form the \glspl{ecm} in Table ~\ref{ECM}.

\begin{table}[htb]
  \caption{The predefined \gls{ecm} configurations, see Table ~\ref{element} for detailed descriptions of each element.}
  \label{ECM}
  \centering
  \begin{tabular}{lcc}
    \toprule
    Name & Number of parameters  & Number of spectra (without filtering) \\
    \midrule
    L-R-RCPE & 5 & 752 (1,084) \\
    L-R-RCPE-RCPE (L-R-2RCPE) & 8 & 1,050 (1,132)\\
    L-R-RCPE-RCPE-RCPE (L-R-3RCPE) & 11 & 1,101 (1,114)\\
    RC-G-G & 6 & 1,099 (1,099)\\
    RC-RC-RCPE-RCPE & 10 & 1,146 (1,152)\\
    RCPE-RCPE (2RCPE) & 6 & 1,060 ( 1,064)\\
    RCPE-RCPE-RCPE (3RCPE) & 9 & 1,129 (1,140)\\
    RCPE-RCPE-RCPE-RCPE (4RCPE) & 12 & 1,134 (1,138)\\
    R-Ws (Rs\_Ws) & 4 & \,\,\, 330 (404)\\
    \bottomrule
  \end{tabular}
\end{table}

\begin{table}[htb]
  \caption{The circuit elements}
  \label{element}
  \centering
  \begin{tabular}{llcc}
    \toprule
    Symbol & Name  & Number of parameters & Equation \\
    \midrule
    L & Inductance & 1 & $j \omega L$ \\
    R & Resistance & 1& $R$ \\
    C & Capacitance & 1 & $\frac{1}{j \omega C}$ \\
    CPE & Constant phase element & 2 & $\frac{1}{C (j \omega)^t}$ \\
    G & Gerischer element & 2 & $\frac{R}{\sqrt{1 + j \omega t}}$ \\
    Ws & Warburg short element & 3 & $R\frac{\text{tanh} \left( (j \omega t)^p \right)}{(j \omega t)^p}$\\
    \bottomrule
  \end{tabular}
\end{table}

\begin{figure}[htb]
    \centering
    \includegraphics[width=.9\linewidth]{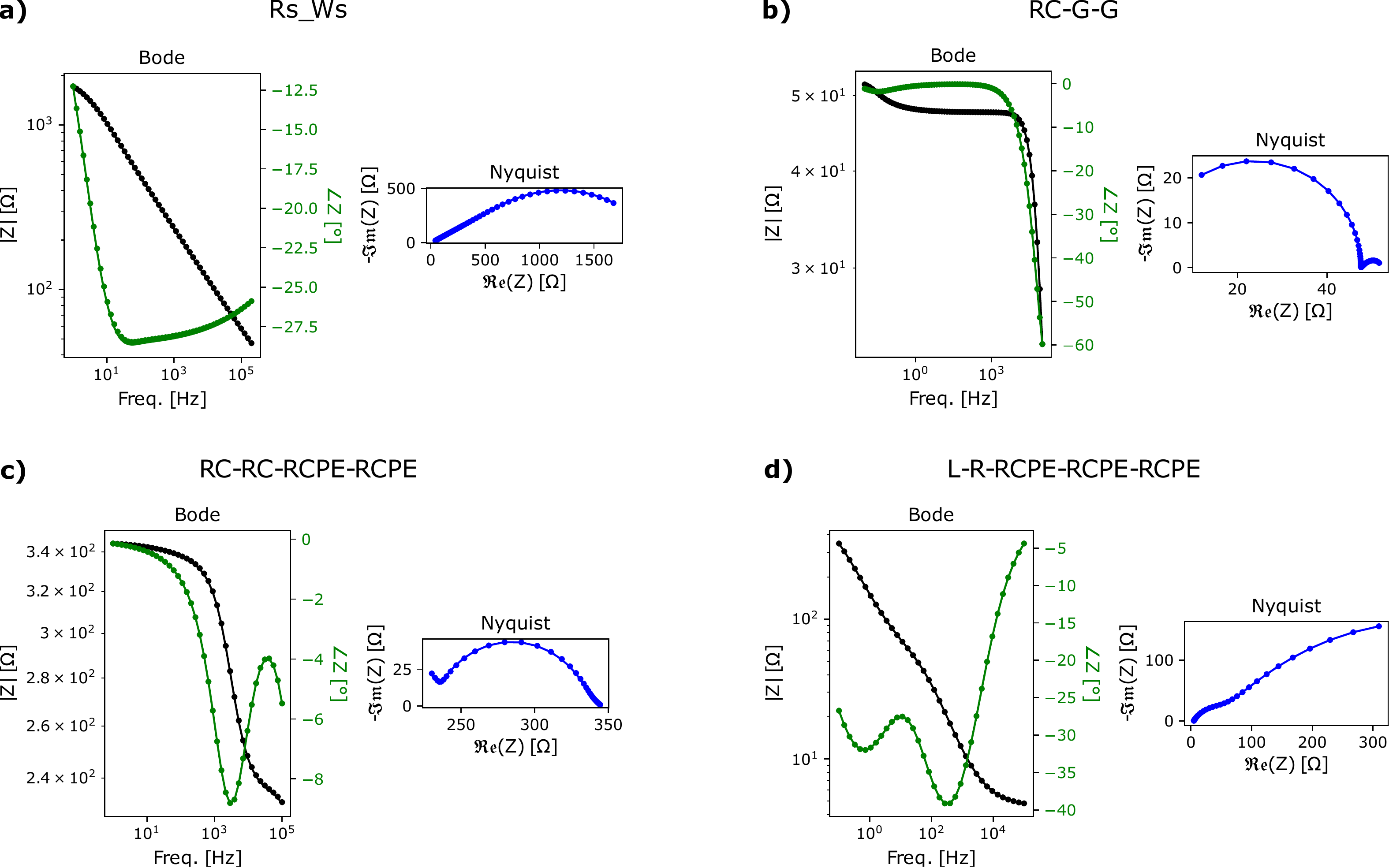}
    \caption{(a)--(d) four selected electrical impedance spectra from the data set. The Bode plot on the left of each subplot shows the magnitude in black and the phase shift in green. The Nyquist plot on the right of each subplot shows the impedance response to each frequency used to excite the system, with each data point in the Nyquist plot corresponding to a distinct frequency. The dots are experimental recordings, and the lines are from equivalent circuit models using circuits and parameters provided by \gls{qs}.}
    \label{fig:data}
\end{figure}

Figure \ref{fig:data} shows four example spectra with different corresponding \glspl{ecm}. The Bode plot on the left of each subplot shows the magnitude in black and the phase shift in green. The Nyquist plot shows the impedance response to each frequency used to excite the system, with each data point in the Nyquist plot corresponding to a distinct frequency; note that a Nyquist plot alone without frequency labeling does not show the data fully, as all detail of the frequency dimension is lost. The values in the Nyquist plot with a small absolute value correspond to the higher frequencies. 

\newpage

\subsection{Data Preprocessing}
\begin{figure}[htb]
    \centering
    \includegraphics[width=.6\linewidth]{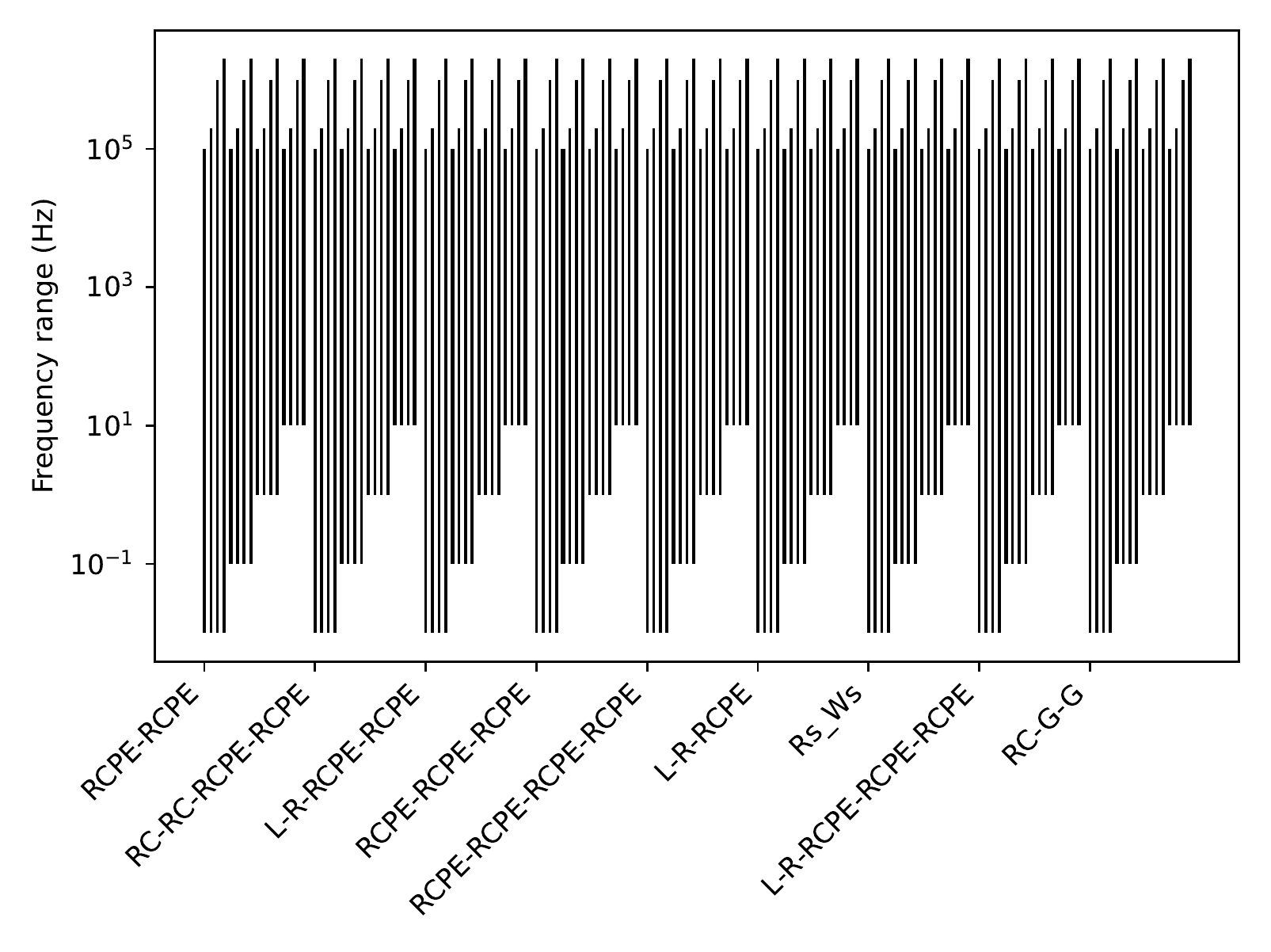}
    \caption{Frequency ranges of the various \gls{ecm} classes.}
    \label{fig:frequency_ranges}
\end{figure}

\begin{figure}[htb]
    \centering
    \includegraphics[width=.6\linewidth]{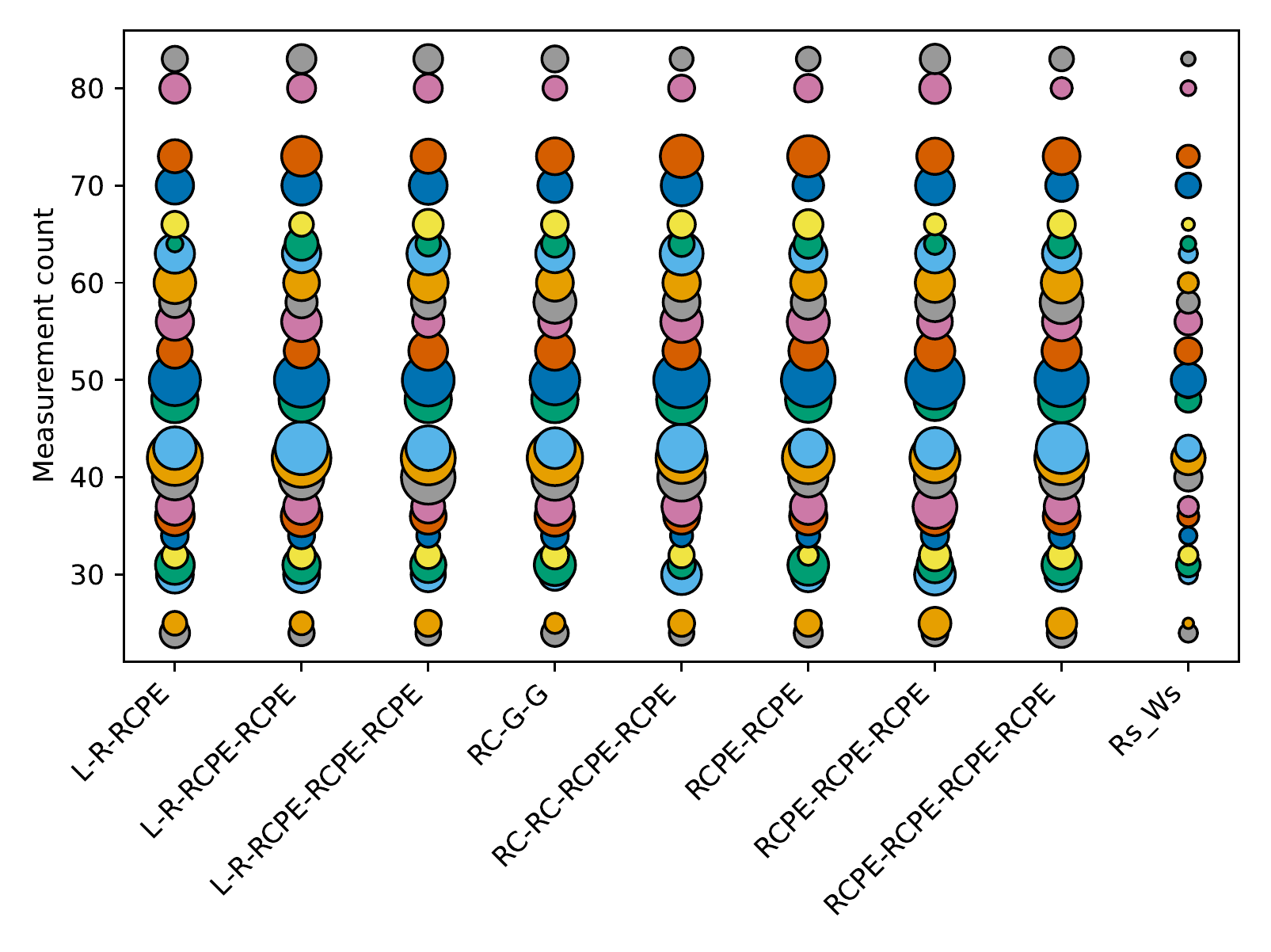}
    \caption{Number of measured frequencies of spectra associated with the various \gls{ecm} classes. The circles are scaled by the number of spectra that have the same number of measured frequencies in the respective \gls{ecm} class.}
    \label{fig:frequency_count}
\end{figure}

The range of frequencies and the number of measured frequencies vary widely between the spectra of each circuit class (c.f., Supplementary Information, Figs.\,\ref{fig:frequency_ranges} and \ref{fig:frequency_count}). The frequency ranges and ranges of the number of measured frequencies are the same for all circuits. Choosing specific frequency ranges is standard practice, depending on the underlying battery's characteristics and the scope of the \gls{eis}-based investigation \cite{FERNANDEZPULIDO20171, krewer2018dynamic}. We interpolated the real and imaginary impedance vectors for every \gls{eis} spectra at $n=30$ logarithmically spaced frequencies across the common frequency basis, ranging from 10$^1$ Hz to 10$^5$ Hz to obtain consistent data for feature design and machine learning. This limited frequency range may neglect certain features, especially at lower frequencies. Another possible approach would be to rescale all frequency ranges to a common vector, though this would lead to further issues since many circuit parameter values are dependent on frequency.

\subsection{Data Visualization}

\begin{figure}[tb]
    \centering
    \includegraphics[width=\linewidth]{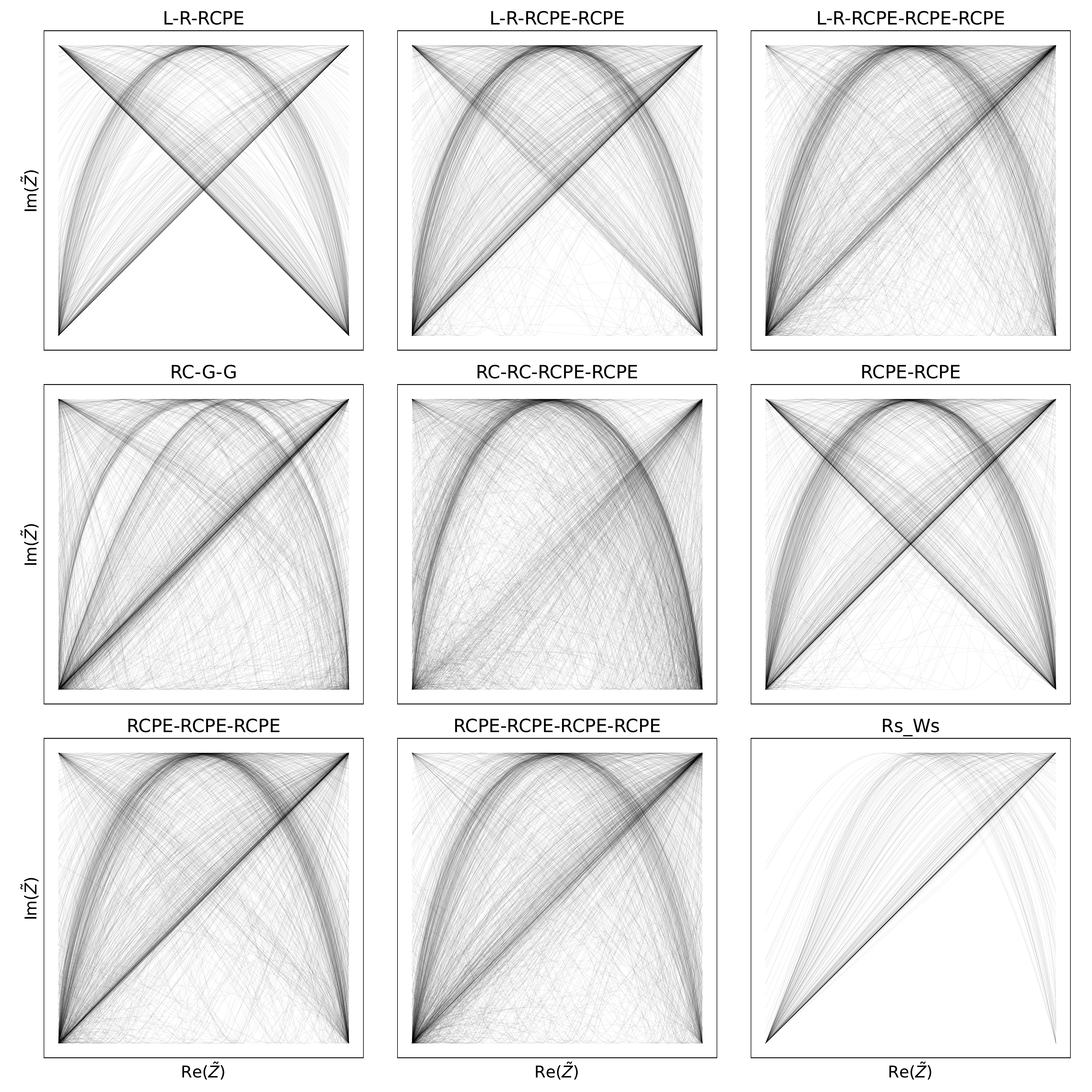}
    \caption{Visualization of normalized impedances of the data set for each circuit class.}
    \label{fig:data_spiderweb}
\end{figure}

Visualization of a large number of impedance spectra is difficult. For this data set, a particular challenge is the distribution of parameters over orders of magnitude, resulting in spectra that also span orders of magnitude. Figure\,\ref{fig:data_spiderweb} visualizes the spectra associated with each circuit class by rescaling them.

\begin{align}
    \operatorname{Re}(\tilde{z_j}) &=  (\operatorname{Re}(z_j) - \min_i{\operatorname{Re}(z_i)})/(\max_i{\operatorname{Re}(z_i)}-\min_i{\operatorname{Re}(z_i)}), \quad j=1,...,n \\ 
    \operatorname{Im}(\tilde{z_j}) &=  (\operatorname{Im}(z_j) - \min_i{\operatorname{Im}(z_i)})/(\max_i{\operatorname{Im}(z_i)}-\min_i{\operatorname{Im}(z_i)}), \quad j=1,...,n
\end{align}

This normalization approach yields $\operatorname{Re}(\tilde{z}_j), \operatorname{Im}(\tilde{z}_j) \in [0, 1]$ but leads to the loss of the information contained in offset, magnitude and magnitude ratio $|\operatorname{Re}(z_i)|/|\operatorname{Im}(z_i)|$. As a consequence, the slope associated with mainly linear spectra is lost, resulting in many spectra that are close to the diagonal of the associated subfigure in Fig.\,\ref{fig:data_spiderweb}, but might look very different when plotted without rescaling.
While Fig.\,\ref{fig:data_spiderweb} does not show any frequency information, it shows the shape of the spectra associated with each circuit. The L-R-RCPE and Rs-Ws circuits have shapes that differ significantly from the other classes. Furthermore, the shapes are increasingly diverse, with an increasing number of RC and RCPE elements. Another critical observation is that a significant overlap of shapes can be observed with semi-circle-shaped and linearly-shaped spectra in all classes, indicating potential classification issues.

Visualizations of the data set in lower dimensions with \gls{umap} are included in the Supplementary Information (\ref{sec:si-vis-dim-red}). The \gls{umap} analysis shows a significant overlap of spectra between different classes. The amount of overlap, however, depends on the classes that are compared. %
The fact that \gls{umap} didn't find a lower dimensional manifold that separates the spectra associated with different \glspl{ecm} well indicates that the supervised learning problem on this data set is challenging, further supporting the intuition gained from Fig.\,\ref{fig:data_spiderweb}.

\section{The Challenge}
\label{sec:challenge_description}
The purpose of the \gls{eis} challenge of the hackathon was to automate the classification of appropriate \glspl{ecm} based on the data set. In particular, the challenge was to create a model that could predict the \gls{ecm} class in the test data set as accurately as possible. In this article, we use the F1-score's weighted average to compare the results of the classification task. We also report the unweighted F1-score's average and the unweighted and weighted average of the recall.

In addition, the automation of guesses for \gls{ecm} parameters was part of the challenge. The guess of the parameters does not need to yield a perfect fit but should be a good initial starting point for more traditional parameter optimization. Due to the difficulty of the classification task, this article focuses entirely on the classification. The Supplementary Information of this article contains more information about parameter estimation and suggests a machine learning approach.

Code for reading the data file, basic \gls{eis} modeling, and plotting of \gls{eis} models versus the recorded data points was provided to the participants to accelerate their efforts. The \gls{eis} model and parameter scoring code are also included in the package so participants can easily self-assess their work. This code can be found in the GitHub repository associated with this article. As well, participants were given examples of how to install the required code environment -- Python, Jupyter, and the SciPy stack -- using \href{https://www.docker.com/}{Docker}, \href{https://anaconda.org/}{Anaconda}, or \href{https://python-poetry.org/}{Poetry}. Jupyter Notebooks containing examples of how to implement the provided code for reading, modeling, plotting, and scoring were provided.

\section{Classification approaches}
\label{sec:approaches}

This section presents the different classification approaches. The models and their performance was studied using a random 80\% train split. The prediction accuracies are subsequently reported on the remaining 20\% of the data that was held out for testing.

\subsection{Random Forest: The baseline model}
Reliable baseline models are essential to quantify performance gains from more complicated models. For example, \gls{lda} and \gls{qda} are linear, static classification methods %
traditionally used for analyzing chemical systems or spectral data. However, these methods do not perform well in classifying \gls{eis} data due to the nonlinearity of the task (cf.\ Fig.\ \ref{fig:umap}). Therefore nonlinear approaches are needed. Here, we present a \gls{rf} model that learns from the raw spectra that were preprocessed to a uniform frequency range according to Section \ref{sec:prepro_freq_range}. Furthermore, the spectral data is arranged in a matrix format:
\begin{equation}
    \mathbf{F}_{\mathrm{Re}} = 
    \begin{bmatrix} 
        \mathrm{Re}(f_1(\omega_1)) & \mathrm{Re}(f_1(\omega_2)) & \cdots & \mathrm{Re}(f_1(\omega_p))\\
        \mathrm{Re}(f_2(\omega_1)) & \mathrm{Re}(f_2(\omega_2)) & \cdots & \mathrm{Re}(f_2(\omega_p))\\
        \vdots  & \vdots & \ddots & \vdots\\
        \mathrm{Re}(f_n(\omega_1)) & \mathrm{Re}(f_n(\omega_2)) & \cdots & \mathrm{Re}(f_n(\omega_p))\\
    \end{bmatrix}
\end{equation}
\begin{equation}
    \mathbf{F}_{\mathrm{Im}} = 
    \begin{bmatrix} 
        \mathrm{Im}(f_1(\omega_1)) & \mathrm{Im}(f_1(\omega_2)) & \cdots & \mathrm{Im}(f_1(\omega_p))\\
        \mathrm{Im}(f_2(\omega_1)) & \mathrm{Im}(f_2(\omega_2)) & \cdots & \mathrm{Im}(f_2(\omega_p))\\
        \vdots  & \vdots & \ddots & \vdots\\
        \mathrm{Im}(f_n(\omega_1)) & \mathrm{Im}(f_n(\omega_2)) & \cdots & \mathrm{Im}(f_n(\omega_p))\\
    \end{bmatrix}
\end{equation}
\begin{equation}
    \mathbf{X} = 
    \begin{bmatrix} 
        \mathbf{F}_{\mathrm{Re}} \ \  \mathbf{F}_{\mathrm{Im}}
    \end{bmatrix} 
\end{equation}
where $f_i(\omega_j)$ denotes the impedance corresponding to the frequency $\omega_j$ of a battery with the index $i$, and Im and Re denote the real and imaginary parts of the impedance, respectively. Each spectrum was scaled by dividing through its maximum real impedance to achieve a consistent scale without decoupling the scale of the real impedance from the scale of the imaginary impedance.

\begin{figure}[htb]
    \centering
    \includegraphics[width=.8\linewidth]{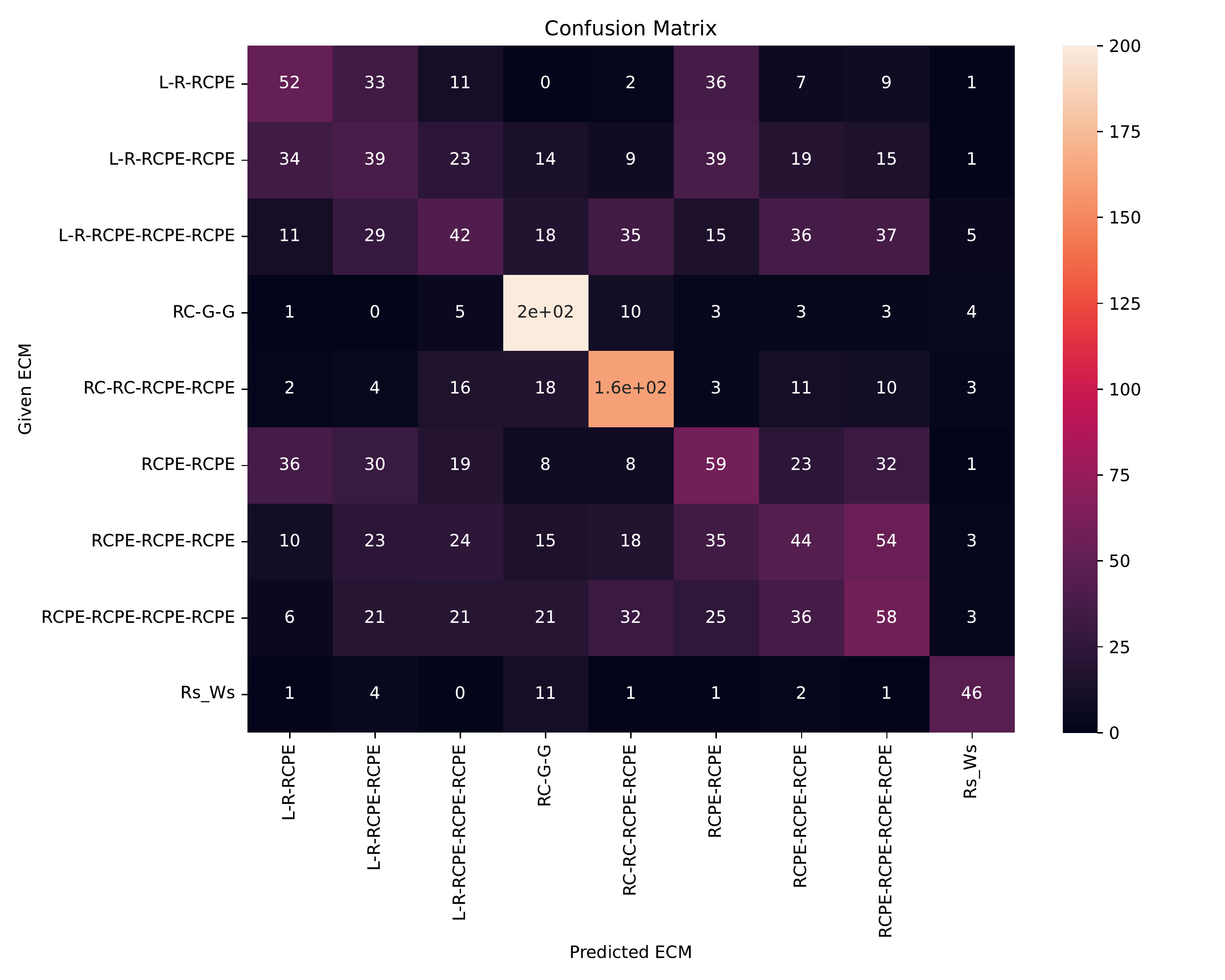}
    \caption{Confusion matrix of predictions on the random 20\% data held out for testing for the \gls{rf} baseline classification model. Weighted F1-score: 0.38}
    \label{fig:rf_confusion_baseline}
\end{figure}

The hyperparameter optimization of the \gls{rf} was carried out with extensive cross-validation. Further details and parameter ranges can be found in the associated code.

The resulting confusion matrix of the test data in Fig.\ \ref{fig:rf_confusion_baseline} showcases that most classes are separated well from one another. However, the model struggles to differentiate between the L-R-RCPE, L-R-2RCPE, and L-R-3RCPE circuits. Similarly, the model struggles to distinguish between the 2RCPE, 3RCPE, and 4RCPE models. These errors are physically sensible, as multiple RCPE elements are often used to fit overlapping peaks. Determining the number of RCPE elements required to accurately fit the data without overfitting is a key challenge when analyzing \gls{eis} data \cite{Buteau_2019}. However, \gls{ecm} types that should be qualitatively much different from one another, such as L-R-nRCPE circuits and R-Ws circuits, are rarely confused.

\subsection{Time-series features, XGBoost: A well-performing solution}
\label{sec:xgb:shap}
The best-performing solution treated each EIS spectrum as a multivariate time series, using the log of frequency as a proxy for time. Classification of time series data often uses engineered features to extract information from the raw data, such as the linearity of the trend, the number of obvious peaks, or the magnitude and phase of periodic fluctuations in the data. To simplify the procedure for proposing possible features and developing methods to extract them, the Python library \textit{tsfresh} was used \cite{christ2018time}. This library extracts hundreds of possible features from time series data and then uses hypothesis testing to remove irrelevant features prior to model training.
The data preprocessing documented in Sec.\ \ref{sec:prepro_freq_range} ensured that all impedance data were evenly spaced with respect to the log of frequency; many of the features generated by \textit{tsfresh} assume that data points are evenly spaced in time. The model architecture used was a gradient-boosted tree implemented by the \gls{xgboost} model architecture, chosen for its high performance in many data science tasks often even without needing substantial hyperparameter optimization or domain expertise \cite{chen2016xgboost}. We optimized the hyperparameters with random search, yielding minor improvements over the model using the default hyperparameters.

\begin{figure}[htb]
    \centering
    \includegraphics[width=.8\linewidth]{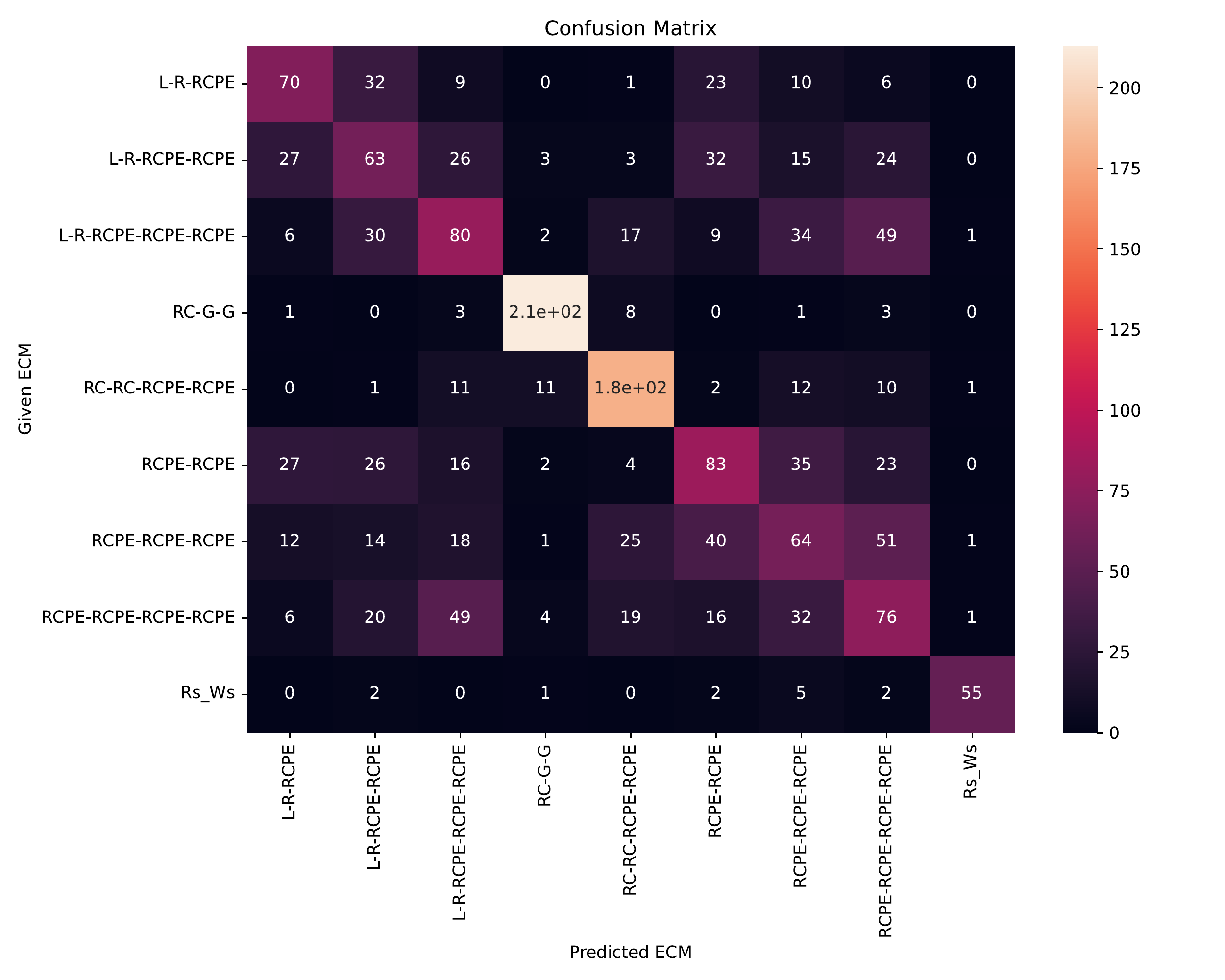}
    \caption{Confusion matrix of predictions on the random 20\% data held out for testing for the tsfresh-XGBoost classification model. Weighted F1-score: 0.50}
    \label{fig:train_confusion}
\end{figure}

Figure \ref{fig:train_confusion} shows the confusion matrix for the predictions on the test set. As noted in the explanation of the data set, there are clear groups of equivalent circuits that are similar to each other, with L-R-nRCPE circuits being easily confused between them but almost never confused with data labeled as R-Ws or RC-G-G circuits, similar to the results of the baseline random forest model but with better performance. The confusion matrix in Fig.\,\ref{fig:train_confusion} shows a clear improvement over the baseline model (Fig.\,\ref{fig:rf_confusion_baseline}), however, qualitatively, the confusion is similar. 

\begin{figure}[htb]
    \centering
    \includegraphics[width=\linewidth]{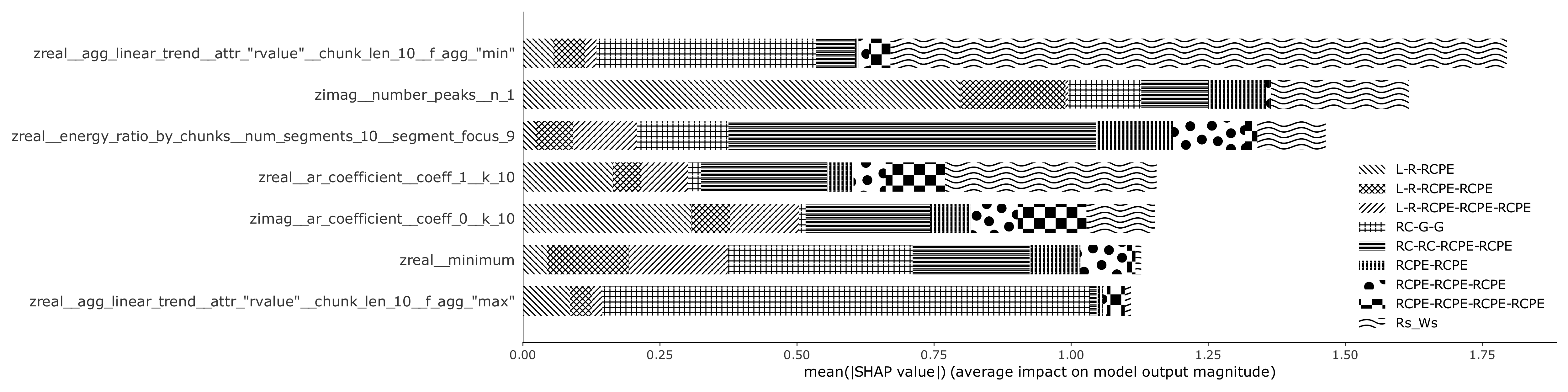}
    \caption{Average feature importance for each class calculated using SHAP on test set predictions for the tsfresh-XGBoost classification model. }
    \label{fig:shap_bar}
\end{figure}

Feature importance for the XGBoost classifier model after generation of time series features using \textit{tsfresh} is investigated using \textit{SHapley Additive exPlanations} (SHAP) \cite{NIPS2017_7062}. Figure \ref{fig:shap_bar} shows the sum of the average SHAP values for all classes for the top 7 features. Of the top 7 features, the importance to each class varies; for instance, the number of peaks in the imaginary impedance has very high importance, on average, for predicting the L-R-RCPE class but almost no importance for predicting the RCPE-RCPE-RCPE class. Another example is the R-value feature of the real impedance (ranked 1st) that has high importance for the diffusion-dominated \gls{ecm} classes (RC-G-G and Rs\_Ws) but little importance to all other classes. This agrees with the physical understanding of diffusion processes generating approximately linear segments in the impedance spectra that are expected to be fitted well by a linear model.
Many of the top 7 features are related to the shape of the impedance spectra. Examples are features related to the linear trend (ranked 1st and 7th) and the number of imaginary impedance peaks (ranked 2nd). 
A key takeaway from the analysis of SHAP values is that there are features that tend to be important for only one or two circuit types (e.g., features ranked 1st, 2nd, 3rd, and 7th), while other features are important for many circuits (e.g., features ranked 4th, 5th, and 6th). However, no single feature alone can classify all spectra accurately. Last, it should be noted that the SHAP values of the features are relatively close. For example, the mean absolute SHAP value of the 7th ranked feature is almost the same as the mean absolute SHAP value of the 4th ranked feature and the top 3 ranked features. Therefore, it is expected that relatively small changes in the data set can lead to shuffled rankings.

\begin{table}[htb]
  \caption{Description of feature in Fig.\,\ref{fig:shap_bar} according to the \textit{tsfresh} documentation \cite{tsfresh_docs}. For more information, we refer to the \textit{tsfresh} documentation that describes the individual functions that calculate the features.}
  \label{tab:tsfresh_feature_explanation}
  \centering
  \scriptsize
  \begin{tabularx}{\textwidth}{>{\hsize=.2\hsize\linewidth=\hsize}X>{\hsize=1.4\hsize\linewidth=\hsize}X>{\hsize=1.4\hsize\linewidth=\hsize}X}
    \toprule
    SHAP Ranking & Feature Name & Description, from \cite{tsfresh_docs}  \\
    \midrule
    1 & zreal\_\_agg\_linear\_trend\_\_attr\_"rvalue"\_\_chunk\_len\_10\_\_f\_agg\_"max" & R-value of linear least-squares regression for values of the time series that were aggregated over chunks versus the sequence from 0 up to the number of chunks minus one, maximum value aggregation\\
    2 & zimag\_\_number\_peaks\_\_n\_1 & Number of peaks of at least support 1 in the time series zimag\\
    3 & zreal\_\_energy\_ratio\_by\_chunks\_\_num\_segments\_10\_\_segment\_focus\_9 & Sum of squares of chunk 9 out of 10 chunks expressed as a ratio with the sum of squares over the whole series zreal\\
    4 & zreal\_\_ar\_coefficient\_\_coeff\_1\_\_k\_10 & First coefficient of an unconditional maximum likelihood of an autoregressive AR(k=10) process.\\
    5 & zreal\_\_ar\_coefficient\_\_coeff\_0\_\_k\_10 & Constant coefficient of an unconditional maximum likelihood of an autoregressive AR(k=10) process.\\
    6 & zreal\_\_minimum & Lowest value of the time series zreal\\
    7 & zreal\_\_agg\_linear\_trend\_\_attr\_"rvalue"\_\_chunk\_len\_10\_\_f\_agg\_"min & R-value of linear least-squares regression for values of the time series that were aggregated over chunks versus the sequence from 0 up to the number of chunks minus one, minimum value aggregation\\
    \bottomrule
  \end{tabularx}
\end{table}

\begin{figure}[htb]
    \centering
    \includegraphics[width=.8\linewidth]{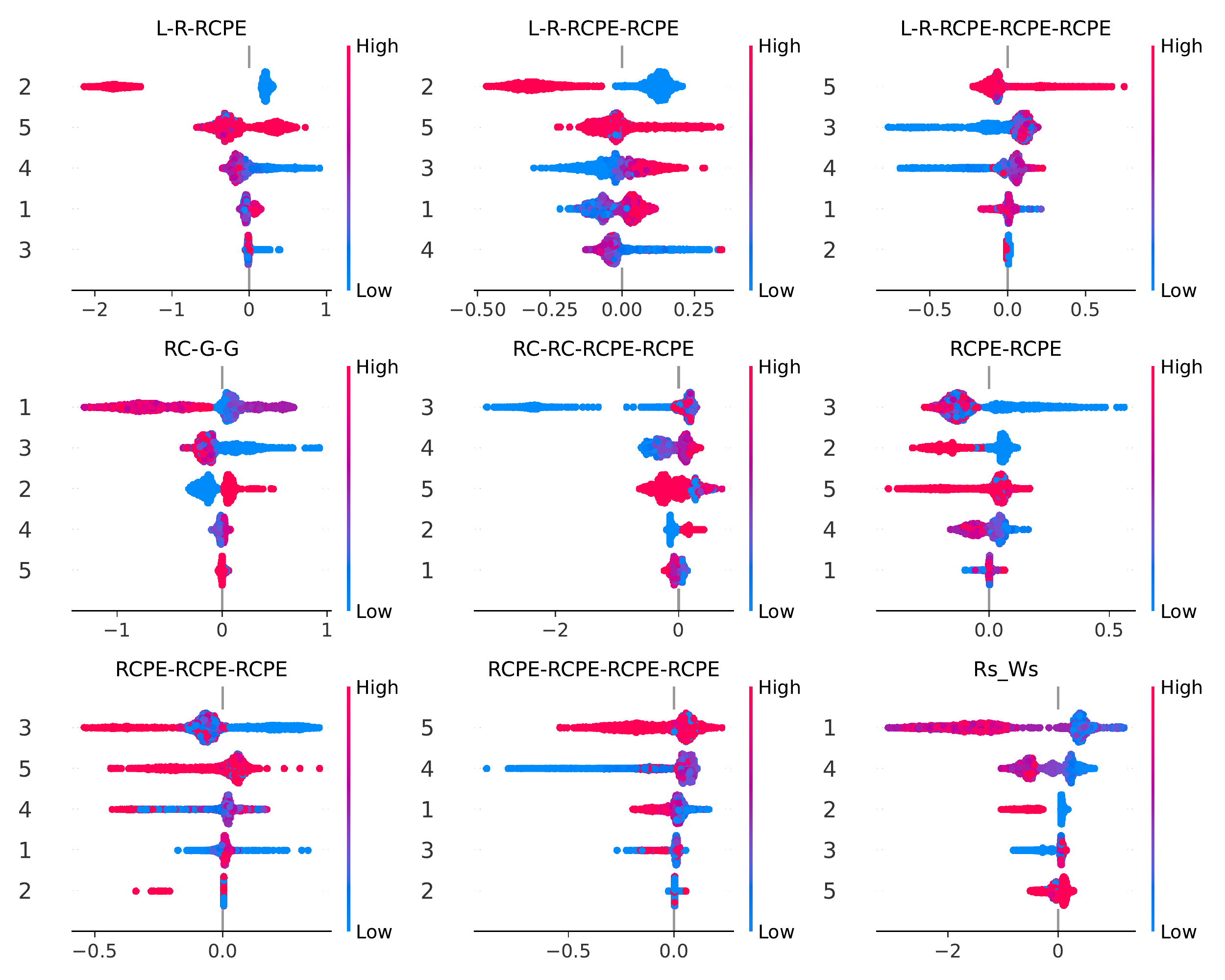}
    \caption{Feature-specific SHAP values from the top 5 average features in Fig.\ \ref{fig:shap_bar} segregated by class for the tsfresh-XGBoost classification model. Each data point corresponds to an observation from the test split. Overlapping points are dispersed to represent the density of values. The x-axis of each plot is the SHAP value for the feature denoted on the y-axis. The features on the y-axis are numbered by their order in Fig.\ \ref{fig:shap_bar}. Points are colored by the value of the feature, with blue corresponding to low values and pink corresponding to high values.}
    \label{fig:shap_class_specific}
\end{figure}

Figure \ref{fig:shap_class_specific} shows SHAP values for every observation by class for each of the top 5 features in Fig.\ \ref{fig:shap_bar}. This figure helps to explain model behavior in more detail. For instance, for the L-R-RCPE circuit, a high value for the number of peaks in the imaginary impedance has a very large negative SHAP value. This reflects domain knowledge, suggesting that impedance spectra with multiple obvious peaks should have more than one RC or RCPE element and thus would not be modeled by the L-R-RCPE circuit. Similarly, Fig.\,\ref{fig:shap_bar} outlined that the RC-G-G and Rs-Ws circuits have a strong dependence on the linearity of the real impedance as mentioned previously. Note that the features plotted here may not necessarily correspond to the most important features for each class on their own; rather, we are just plotting the top 5 average features to simplify comparison.

\subsection{Convolutional Neural Network: The creative approach}

\glspl{cnn} are commonly used to analyze image data and can learn complex features and relationships. In the past, efforts using artificial neural networks to analyze large amounts of \gls{eis} data, without humans having to choose initial parameter values for the equivalent circuits, have been employed. Buteau and Dahn used an inverse model parameterized with a convolutional neural network over a data set containing 100,000 impedance spectra \cite{Buteau_2019}. Rastegarpanah et al.\ \cite{doi:10.1177/0959651820953254} developed a rapid neural network starting with a single hidden layer baseline model, optimized by a Gaussian process hyperparameter scheme, to estimate the state of health of Nissan Leaf 2011 battery modules using a data set of 106 samples. 

Here we present a \gls{cnn} model for classifying \glspl{ecm}. This approach is motivated by the fact that experts also look at impedance spectra visually because the shape of a spectrum is essential to determine an appropriate \gls{ecm}. Furthermore, the SHAP feature analysis in Sec.\,\ref{sec:xgb:shap} showed that features related to the shape of the impedance spectra are important for \gls{ecm} classification. The preprocessed impedance spectra were visualized in a Nyquist plot. The x-axis  corresponds to the real part and the y-axis to the imaginary part of the impedance.
\begin{figure}[htb]
    \centering
    \includegraphics[width=.8\linewidth]{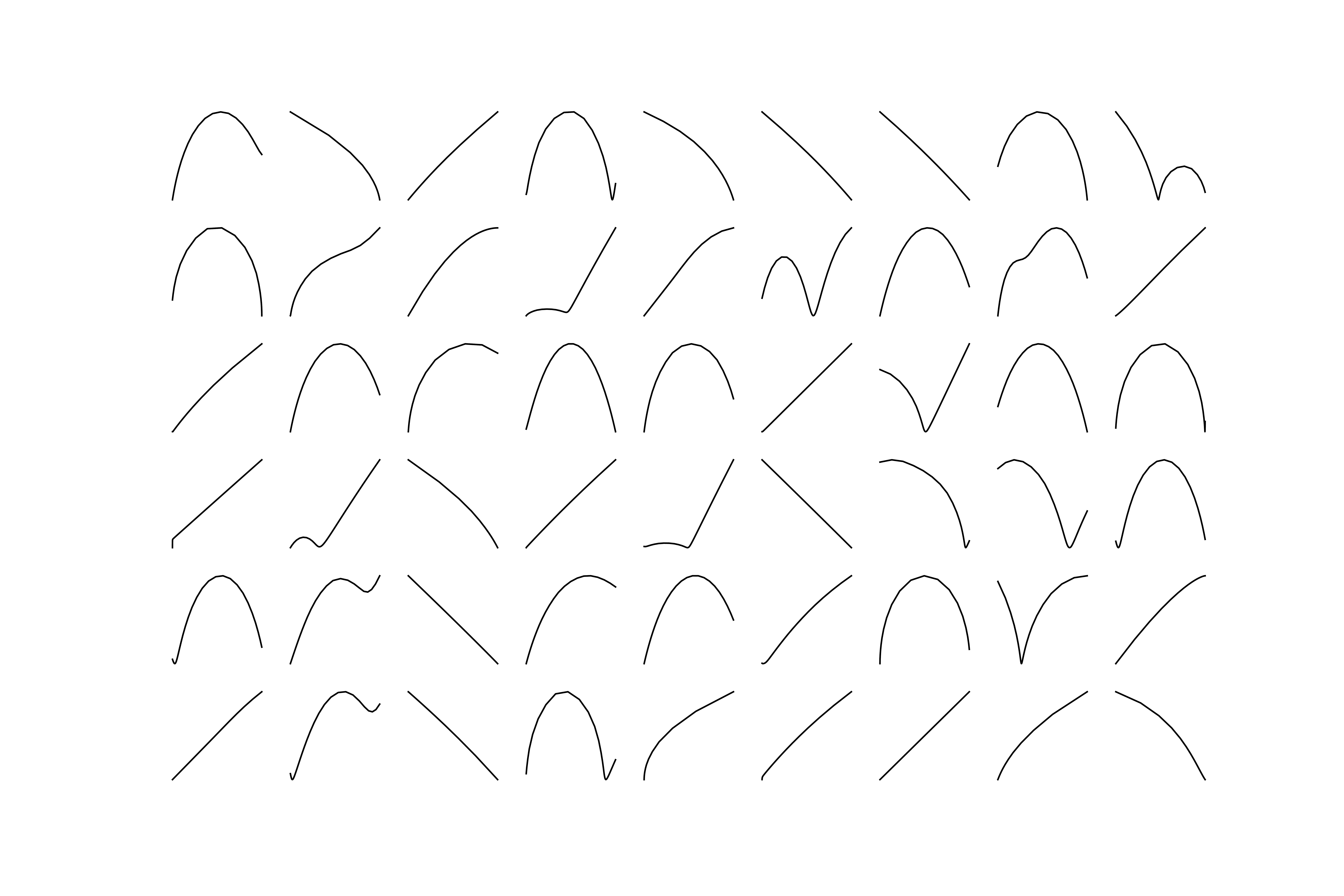}
    \caption{Different \gls{eis} spectra interpolated to a common frequency range visualized in a Nyquist plot.}
    \label{fig:eis_art}
\end{figure}
Figure \ref{fig:eis_art} shows a small subset of the generated images. Each spectrum can be interpreted as a battery's signature. While many spectra look very different, similar patterns can be identified. There are similarities between the \gls{eis} images and the MNIST digit classification data set for which many proven, easy-to-use architectures exist. For a proof of concept \gls{cnn}, we decided on a simple network that yielded a high performance on the MNIST data set \cite{kaggle_notebook}. We modified the kernel and stride sizes of the first layer to account for the image resolution and structure. To keep training time low, we used a resolution of 56$\times$56 pixels. 

\begin{figure}[htb]
    \centering
  \includegraphics[width=0.78\linewidth]{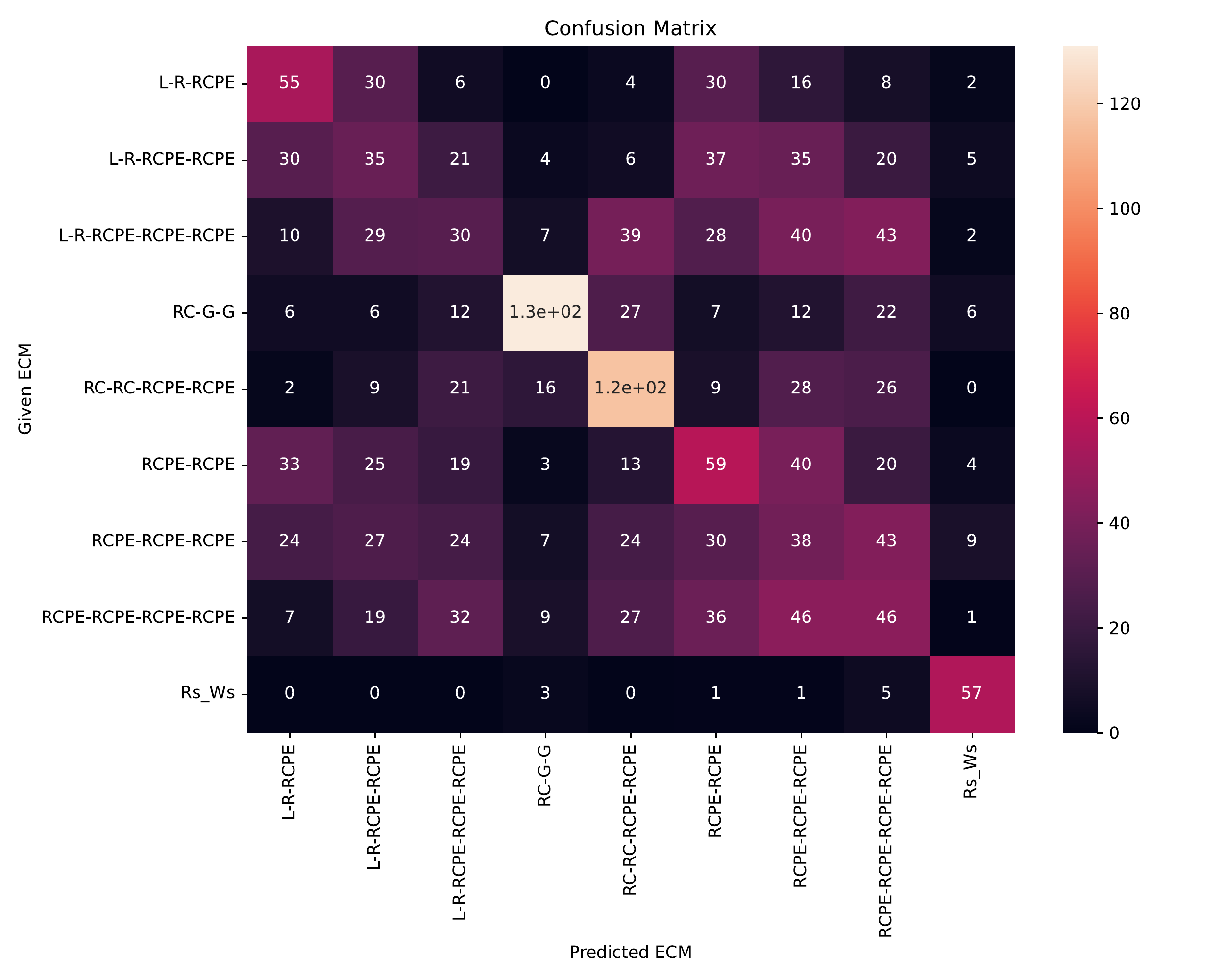}
    \caption{Confusion matrix of predictions on the random 20\% data held out for testing for the \gls{cnn} classification model. Weighted F1-score: 0.32}
    \label{fig:cm_cnn_class}
\end{figure}

The confusion matrix shown in Fig.\ \ref{fig:cm_cnn_class} shows a similar pattern to Figs.\,\ref{fig:rf_confusion_baseline} and  \ref{fig:train_confusion}. The F1-score of 0.33 is lower than for the other models. Nevertheless, this result is a promising proof of concept, given that the network was only slightly adapted, and the frequency and magnitude information is not considered. The same \gls{cnn} architecture was tested using images with colored lines encoding the frequency information. However, the F1-score improvements were not statistically significant. This suggests that the model with the tested architecture cannot account for the frequency information. How to incorporate the frequency and magnitude information into the model and how to design suitable \gls{cnn} architectures for classifying \glspl{ecm} remain open questions.

\paragraph{Lessons from a Transfer Learning Approach} Transfer learning in the context of (deep) neural networks refers to using a network with a defined architecture that was trained on one data set for another somehow related application. The idea is that the embeddings learned by the network will also be helpful for the new task. During the BatteryDEV hackathon, a transfer learning approach based on the MobileNetV2 architecture was suggested, see Supplementary Information (\ref{sec:si:sup_info_cnn_freqColored}).
However, this approach was not (yet) successful. The computational costs to handle a deep network like MobileNetV2 are high. Consequently, it is costly to experiment and tune the model. Furthermore, the original training data of the MobileNetV2 were color images of objects, and the resulting embeddings generated by the layers close to the final layer of the network, do not work well to classify \gls{eis} spectra. One explanation is that the statistics of (natural) images follow special distributions \cite{ruderman1994statistics}, which are very different from the sparse nature of the \gls{eis} data. Consequently, the embeddings learned from the training data of the MobileNetV2 do not suit the \gls{eis} classification task. A possible solution for this issue would be to allow for retraining the parameters of the MobileNetV2. However, this is non-trivial given the small amount of fewer than 10k spectra and was thus not pursued further. 

\subsection{Comparison of Classification Results}

The classification results show that the highest prediction accuracy was obtained by the tsfresh-XGBoost approach, clearly outperforming the other two models. However, the CNN approach was only investigated as a proof of concept, and there is still potential for further performance improvements. 
\begin{table}[htb]
  \caption{Classification results, (results for unfiltered data). The reported accuracies for the \gls{cnn} are average accouracies of ten trained networks.}
  \label{tab:class_res_overview}
  \centering
  \begin{tabular}{lcccc}
    \toprule
    Approach & F1-score, macro & F1-score, weighted & Recall, macro & Recall, weighted\\
    \midrule
    RF Baseline & 0.40 (0.43) & 0.38 (0.41) & 0.41 (0.45) & 0.40 (0.42)\\
    tsfresh-XGBoost & \textbf{0.52 (0.54)} & \textbf{0.50 (0.52)} & \textbf{0.52 (0.54)} & \textbf{0.50 (0.52)}\\
    CNN & 0.35 (0.36) & 0.32 (0.33) & 0.36 (0.37) & 0.32 (0.34)\\
    \bottomrule
  \end{tabular}
\end{table}
Furthermore, Tab.\,\ref{tab:class_res_overview} shows that the models trained on the entire data set without filtering out any spectra performed slightly better than their counterparts that were only trained on the data that passed the filter criteria. First, many removed spectra were associated with the L-R-RCPE and Rs-Ws circuits (c.f.\, Tab.\ref{ECM}). In addition, the filtered-out spectra are quite different from the other spectra. The models based on the entire data thus managed to classify them slightly better on average than the other spectra. However, the filtered-out spectra are mostly unphysical and unlikely to be observed from a real battery. Thus, the models based on the unlabeled data set picked up relationships that are not expected to generalize beyond this data set. 

\begin{figure}[htb]
    \centering
    \includegraphics[width=\linewidth]{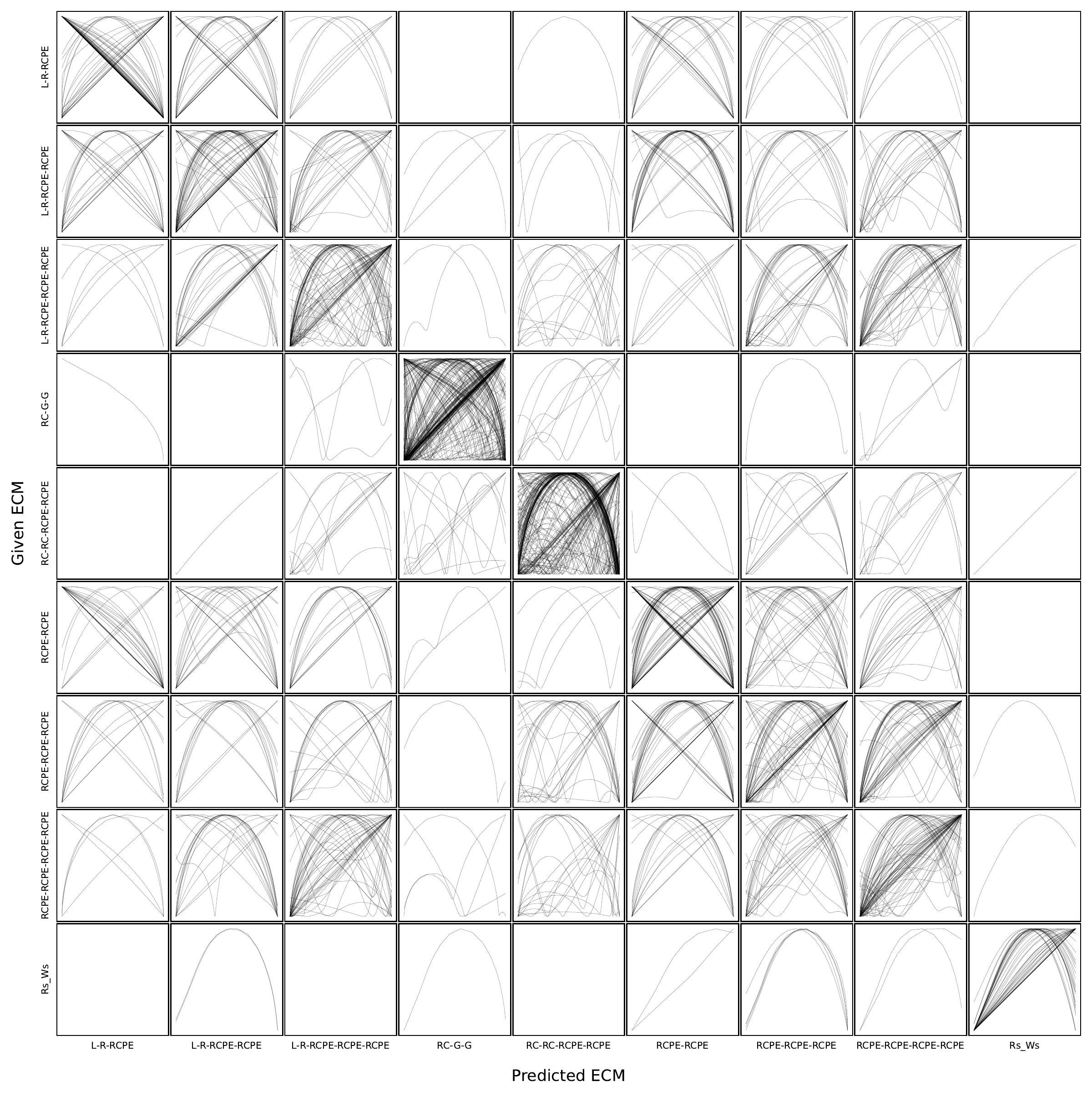}
    \caption{Confusion matrix of normalized impedance spectra associated with the XGB-tsfresh model.}
    \label{fig:cm_nyquist}
\end{figure}

Although very different, the three investigated approaches share similar patterns in their confusion matrices (compare Figs. \ref{fig:rf_confusion_baseline}, \ref{fig:train_confusion}, and \ref{fig:cm_cnn_class}). All approaches struggle to distinguish the L-R-RCPE, L-R-2RCPE, and L-R-3RCPE circuits. Similarly, the classification methods struggle to distinguish the 2RCPE, 3RCPE, and 4RCPE circuits. Furthermore, there is high confusion between the L-R-3RCPE and 4RCPE circuit classes. The other combinations of L-R-nRCPE and nRCPE circuits also show higher levels of confusion relative to the combinations not mentioned here. A potential explanation is that the above-mentioned \glspl{ecm} can generate similar spectra. Figure\,\ref{fig:cm_nyquist} supports this claim because many misclassified spectra show patterns that also agree well with other circuit types. For example, the spectra classified as L-R-RCPE but associated with RCPE-RCPE and those classified as RCPE-RCPE but associated with L-R-RCPE look similar to a human observer. Also, many of the misclassified spectra have a semi-circle shape which can be generated by different \glspl{ecm}, hinting at potential identifiability issues. A key observation is that the patterns shared by the columns of Fig.\,\ref{fig:cm_nyquist} (i.e., the spectra that were classified to be associated with the same circuit) appear to be more similar than the patterns shared by the rows (i.e., the spectra that are associated with the same circuit). However, Fig.\,\ref{fig:cm_nyquist} suffers from the same drawbacks as Fig.\,\ref{fig:data_spiderweb} (i.e., loss of magnitude, magnitude ratio, and frequency information).

\section{Discussion}
\label{sec:discussion}

While broad, this \gls{eis} data set does not represent the whole variety of impedance measurements observed from \gls{lib}. The fact that the underlying \gls{ecm} parameters were drawn from independent distributions results in a data set containing spectra that are unlikely to be observed by a real battery or unphysical. While we addressed this drawback by filtering out spectra, we suspect spectra remain in the data set that would be unlikely to be observed during the R\&D of new battery materials. Furthermore, certain parameter combinations of models with many parameters might generate spectra that could likely be modeled by simpler \glspl{ecm}.

The choice of \gls{ecm} classes and parameter bounds was informed by the R\&D of solid-state batteries. The electrode materials generally used by solid-state and liquid-electrolyte batteries are similar; however, solid electrolytes' diffusivity measurements include grain-boundary and bulk diffusion effects, while liquid electrolytes have no grain-boundary effects, which can lead to differences in impedance spectra. Given the complexity of the systems to which \glspl{ecm} modeling is typically applied, automating the parameterization of \glspl{ecm} would be helpful to automate \gls{ecm} modeling for non-experts or to process big data of impedance spectra. However, it is uncertain how well the trained models presented in this article would perform on different data sets, e.g., measured spectra from solid-state batteries or measured spectra from liquid electrolyte batteries. Nevertheless, the proposed approaches are flexible and can be used to learn from other \gls{eis} data sets.

A further limitation of this data set is that it limits the supervised learning problem to predicting one of the nine possible \gls{ecm} types. A more generally useful result would be the generation of candidate \glspl{ecm} in an unsupervised learning approach, which could then learn from data sets such as this to then propose useful \glspl{ecm} on new systems. Furthermore, the model selection and its parameter estimation are preferably performed in one go, as the motivation of \gls{ecm} analysis is to parameterize and quantify the raw impedance spectra to comparable variables. Although a classification suggests which model to select, it does not provide the parameter values or initial guess for its fitting process.

To this end, the second stage of the BatteryDEV hackathon was supposed to be the parameter estimation of \glspl{ecm}. Still, the limited duration of the hackathon hindered the complete exploration of this direction. The parameterization was defined to estimate best-fit circuit parameters. Preliminary results of the regression task with the tsfresh-XGBoost approach and comparison to the parameters provided by \gls{qs} are included in the Supplementary Information (\ref{sec:si:reg}). Furthermore, a Bayesian inference approach was conceptualized; see Supplementary Information (\ref{sec:si:bayes}).

\section{Conclusion}
\label{sec:conclusion}
The \gls{eis} classification challenge led to the exploration and development of novel \gls{ecm} classification methods. The presented approaches cut the time engineers spend on model selection for electrochemical impedance spectra, allowing them to focus on modeling and making better conclusions. The best-performing model used the \textit{tsfresh} library to automatically calculate features of the impedance spectra and an \gls{xgboost} model for the classification of \gls{eis} spectra into the appropriate \gls{ecm} class. The \gls{rf} model based on the raw spectral data performed slightly worse. The \gls{cnn} approach showed that \glspl{cnn} can classify \gls{eis} spectra. However, challenges remain to apply \glspl{cnn} to accurately classify impedance spectra. Future studies can refine and build upon the techniques and benchmarks described in this article.

With this article, we publish the analyzed data set of 9,300 impedance spectra provided by \gls{qs}. The software and data for this article are available as open source in the corresponding GitHub repository. This work demonstrates how companies can contribute to and leverage open-source innovation. We hope that work will pave the way for more such collaborations.

\section*{Data and Code Availability}
\label{sec:data_code_avail}
The data and code are available on the corresponding GitHub repository: \url{https://github.com/BatteryDEV/AutoECM}. QuantumScape (QS) provided the EIS data contained in the repository. The first data set comprises approximately 9,300 synthetic spectra with the associated Equivalent Circuit Model (ECM). The second data set contains approximately 19,000 unlabeled spectra consisting of about $80\%$ synthetic and $20\%$ measured data. The parameter ranges for all synthetic data are informed by the R\&D of \gls{qs}. The measured spectra are from a range of different materials, with some replicate measurements at different temperatures, and/or \gls{soc}, and/or \gls{soh}. The code comes with an open-source MIT license, and the data are available openly under the terms of the CC BY license. 

\section*{Author Contributions}
Joachim Schaeffer: Formal Analysis, Funding acquisition, Methodology, Project administration, Supervision, Software, Writing; Paul Gasper: Formal Analysis, Methodology, Software, Writing; Masaki Adachi: Funding acquisition, Project administration, Software, Writing; Raymond Gasper: Funding acquisition, Project Administration, Software; Esteban Garcia-Tamayo: Formal Analysis, Methodology, Writing; Juan Pablo Gaviria-Cardona: Formal Analysis, Methodology, Software; Simon Montoya-Bedoya: Formal Analysis, Methodology, Writing, Software; Richard D. Braatz: Writing – review \& editing; Rolf Findeisen: Writing – review \& editing; Anoushka Bhutani: Funding acquisition, Project Administration, Software, Writing; Andrew Schiek: Software, Formal Analysis, Writing; Rhys Goodall: Software, Formal Analysis, Writing – review \& editing; Simon Engelke: Funding acquisition, Project administration, Supervision, Writing. 

\section*{Acknowledgments}
The first BatteryDEV hackathon was supported through the 10toGO hackathon initiative by Volkswagen and Microsoft. The second BatteryDEV hackathon had sponsors from industries spanning consumer electronics, power tools, utilities, HR, recycling, startups, and academia.
We thank QuantumScape for making their EIS data set available; in particular, Tim Holme for supporting the BatteryDEV initiative and the effort to make these data publicly available. Furthermore, we thank Ryan Lu, Krishna Ayer, and Michael Plews for their assistance during the office hours of the hackathon.
We thank the Battery Associates team for supporting the hackathon.

Parts of the work for this paper were done during Joachim Schaeffer's time at the Massachusetts Institute of Technology.

Paul Gasper, Rhys Goodall, and Andrew Schiek thank the efforts and time for the rest of 'Team Battmen' from the BatteryDev 2022 competition: Tushar Deshai, and Hugo Leduc.

Likewise, appreciation for the time and effort from the rest of BatteryDev2022's runner-up 'Team ejjn': Juan E. Betancur,  Juan P. Tamayo, Nicolas Montoya-Escobar and Michael Guzman-De Las Salas.

\section*{Funding}
Financial support for Joachim Schaeffer's time at the Massachusetts Institute of Technology is acknowledged by a fellowship within the IFI program of the German Academic Exchange Service (DAAD), funded by the Federal Ministry of Education and Research (BMBF).

Paul Gasper and Andrew Schiek are supported by the Assistant Secretary for Energy Efficiency and Renewable Energy, Office of Vehicle Technologies of the U.S. Department of Energy (DOE) through the Machine Learning for Accelerated Life Prediction \& Cell Design program, technology manager Dr. Simon Thompson. The National Renewable Energy Laboratory is operated by Alliance for Sustainable Energy under Contract No.\ DE-AC36-08GO28308 for the U.S. Department of Energy. The views expressed in the article do not necessarily represent the views of the DOE or the U.S. Government. The U.S. Government retains and the publisher, by accepting the article for publication, acknowledges that the U.S. Government retains a nonexclusive, paid-up, irrevocable, worldwide license to publish or reproduce the published form of this work, or allow others to do so, for U.S. Government purposes.

\section*{Competing Interests}
Masaki Adachi is an employee of Toyota Motor Corporation and is the founder of Inferable Energy O{\"U} and is affiliated with the University of Oxford. Rhys Goodall is an employee of Chemix.ai. Simon Engelke is the founder of Battery Associates. Esteban Garcia-Tamayo is an employee of Titan Advanced Energy Solutions.

\section*{Inclusion and Diversity} 
We support inclusive, diverse, and equitable conduct of research.

\printbibliography
\clearpage
\begin{refsection}
\setcounter{page}{1}
\counterwithin{figure}{subsection}
\counterwithin{table}{subsection}
\renewcommand{\thesubsection}{\Alph{subsection}}
\section*{Supplementary Information for ``Machine learning benchmarks for the classification of equivalent circuit models from electrochemical impedance spectra''}
\label{sec:si:sup_info}

\subsection{Data Preprocessing}
\label{sec-si-dapre}

\subsubsection{Filtering}
\label{sec-si-dapre-filter}
The provided synthetic data set contains spectra that are unphysical or very unlikely to represent a physical battery or both. Such spectra are part of the data set because the generating equivalent circuit models rely on parameters sampled from independent reciprocal distributions as described in Sec.\,\ref{sec:data}. 

The labeled data set, $\mathcal{Z}$, contains 9327 individual spectra $\mathbf{z} \in \mathcal{Z}$ that are vectors of impedances, $\mathbf{z} \in \mathbb{C}^{n}$.

\begin{figure}[H]
    \centering
    \includegraphics[trim={0 6.8cm 0 0},clip, width=.99\linewidth]{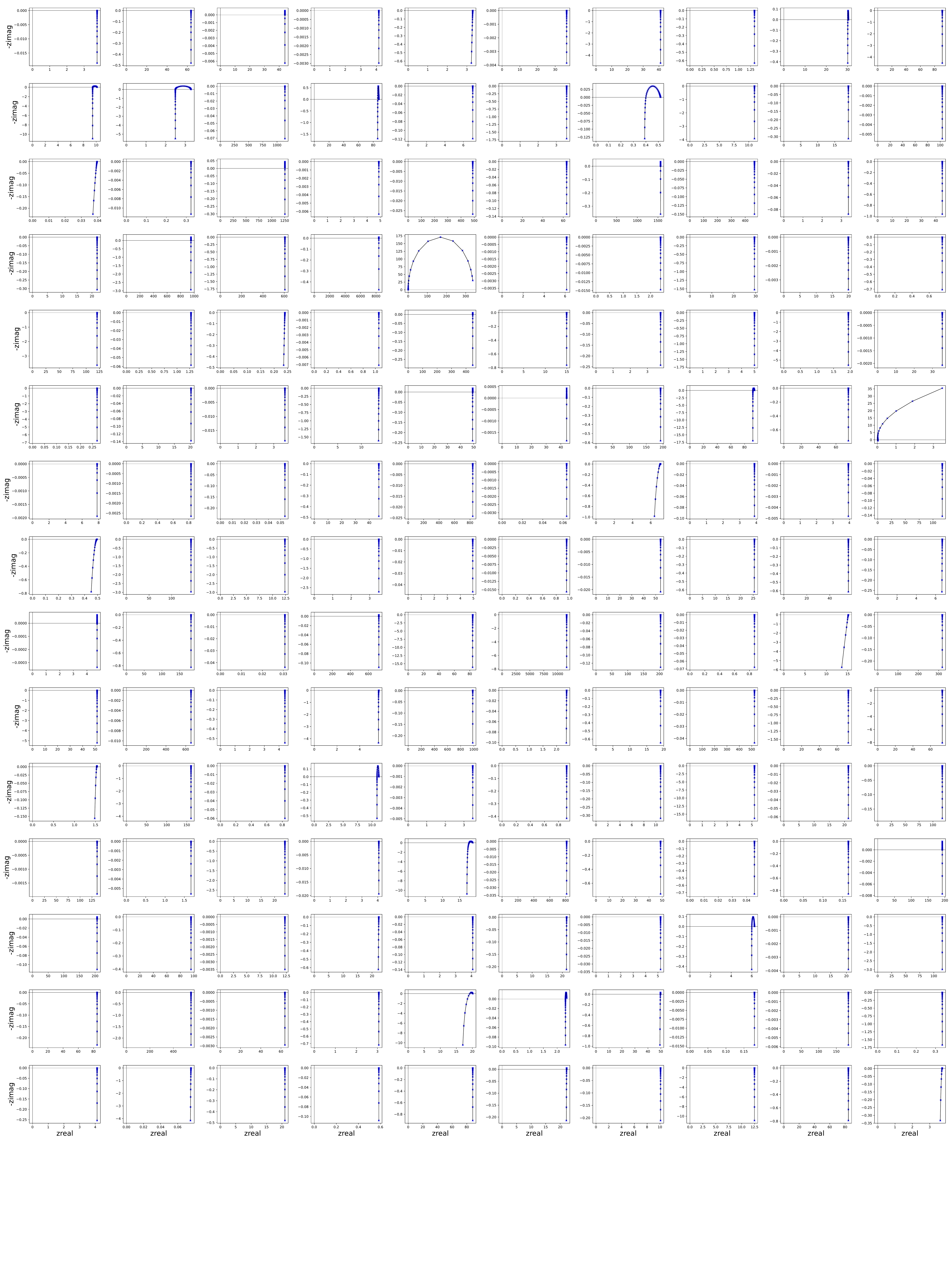}
    \caption{Impedance spectra associated with the L-RCPE circuit class that were filtered out based on criteria described in Tab.\,\ref{tab:filter_criteria}, 150 spectra were randomly sampled from the 332 spectra that were filtered out in total.}
    \label{fig:filtered_out_L-RCPE}
\end{figure}

\begin{table}[htb]
  \caption{Filtering Criteria}
  \label{tab:filter_criteria}
  \centering
  \begin{tabular}{l|l}
    \toprule
    \multicolumn{2}{l}{1. Remove all impedances $z_i$ from each spectrum $\mathbf{z}$ with: }\\
    \midrule
    $\Im(z_i) > 0$ & Positive imag. impedances \\ 
    \midrule
    \multicolumn{2}{l}{2. Remove all spectra $\mathbf{z}$ that have:} \\
    \midrule
    $|\set{z_i | \Im(z_i) < 0}|  < 10 $ & Less than 10 impedances with negative imag. part \\
    \midrule
    $|\max_i (-z_i) | < 0.5 |\min_i (-z_i)|$ & Negative img. range smaller than half the positive imag. range \\
    \midrule
    $\exists\,i: \Re(z_i) < 0$ & Negative real impedances\\
    \midrule
    $\exists\,i,j: i>j$ and $(\Re(z_i) - \Re(z_j)) > a$  & Not \textit{"mainly"} decreasing real impedance with increasing frequency \\
    $a = -0.03$\\
    \bottomrule
  \end{tabular}
\end{table}
The filter criteria removed 422 spectra from 9327 spectra in total, thereby removing approximately $4.5\%$ of the data. Figure \ref{fig:filtered_out_L-RCPE} shows the spectra associated with the L-RCPE circuit that were filtered out. Most of the spectra in Fig\,\ref{fig:filtered_out_L-RCPE} are mainly in the fourth quadrant of the zreal, -zimag plane or are close to vertical lines or both. 

\begin{figure}[H]
    \centering
    \includegraphics[trim={0 6.8cm 0 0},clip, width=.99\linewidth]{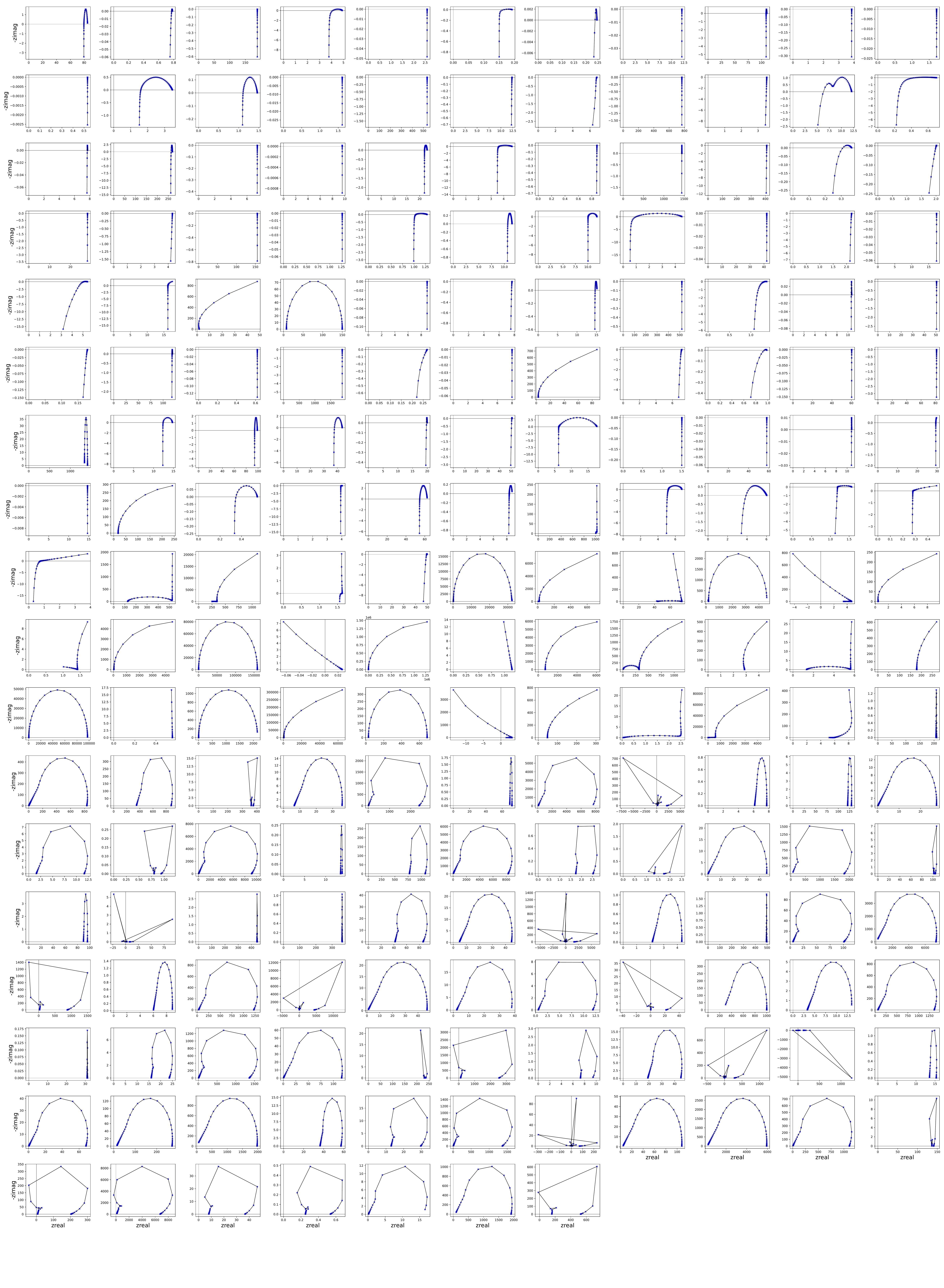}
    \caption{Impedance spectra associated with the denoted \gls{ecm} that were filtered out based on criteria described in Tab.\,\ref{tab:filter_criteria}. The spectra are ordered by \gls{ecm} type. L-R-RCPE-RCPE (82 spectra), L-R-RCPE-RCPE-RCPE (13), RC-RC-RCPE-RCPE: 6 spectra), RCPE-RCPE (4 spectra), RCPE-RCPE-RCPE (11 spectra), RCPE-RCPE-RCPE-RCPE (4 spectra), Rs-Ws (74 spectra).}
    \label{fig:filtered_out_other}
\end{figure}

Figure \ref{fig:filtered_out_other} shows the filtered-out spectra associated with the other \glspl{ecm} circuit types, sorted by circuit type. The bottom seven rows contain only spectra related to the Rs-Ws circuit. Most of these spectra are unphysical, with impedances corresponding to neighboring frequency values changing wildly and negative real impedances or both.

\subsubsection{Frequency Range and Number of Data Points}
\label{sec:prepro_freq_range}

\subsection{Visualization of Nonlinear Dimensionality Reduction}
\label{sec:si-vis-dim-red}
\begin{figure}[htb]
    \centering
    \includegraphics[width=.8\linewidth]{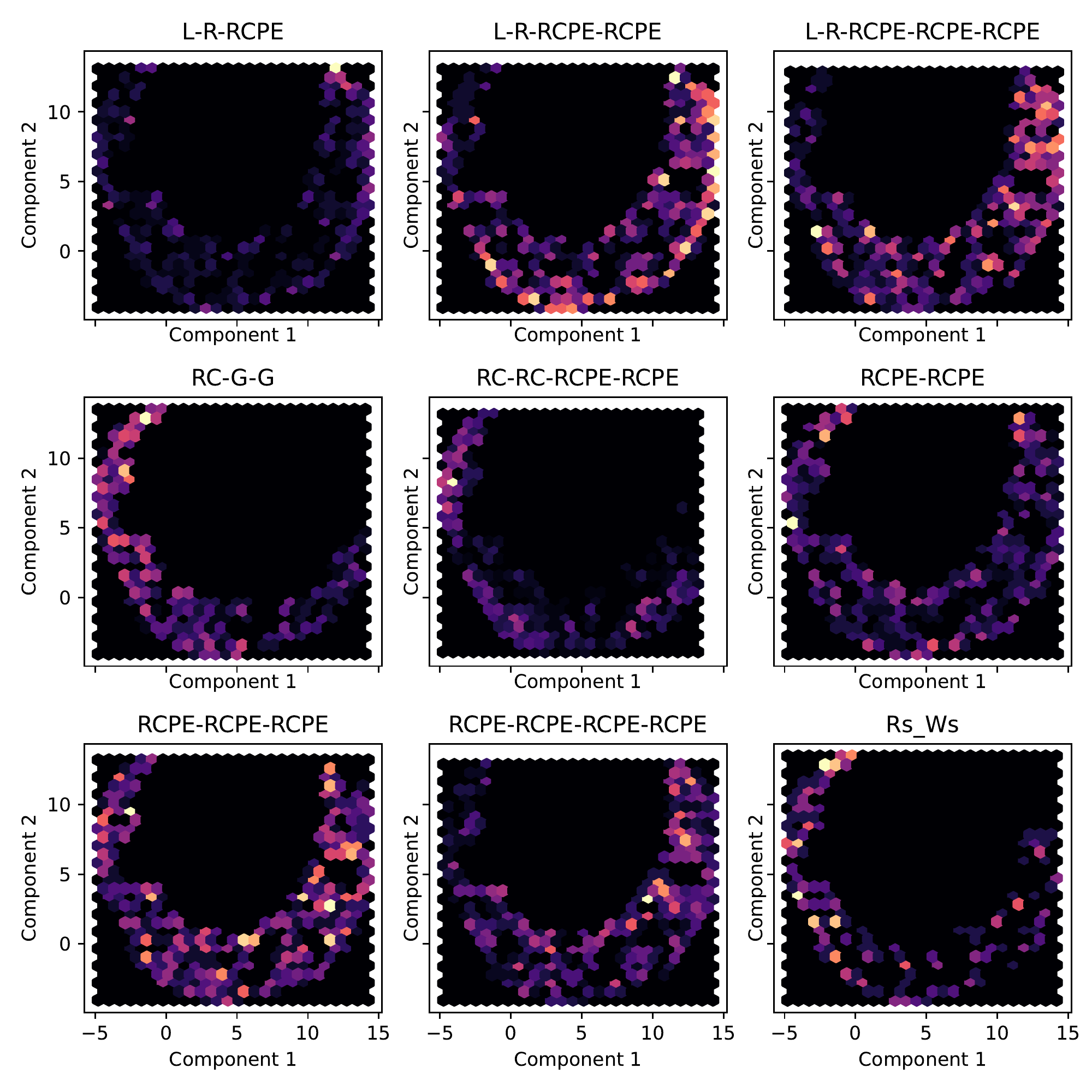}
    \caption{Two-component \gls{umap} reduction of the data set. Because clusters are substantially overlapping, component values for each circuit type in the data set are plotted on separate axes. To help distinguish the density of points, each axis is plotted as a binned hexagonal grid, where bright regions correspond to high bin counts and dark regions correspond to low bin counts. \gls{umap} is conducted after data frequency range preprocessing.}
    \label{fig:umap}
\end{figure}

Data visualization through dimensionality reduction helps to understand the data's structure which can be challenging to observe directly. Figure \ref{fig:umap} shows the entire data set after dimensionality reduction to just two components using \gls{umap} \cite{mcinnes2018umap} to empower visualization of the entire data set on a single axis. Because the components for each circuit type substantially overlap, the components for each circuit type are plotted as a binned hexagonal grid. Overall, the color-coded shape of the reduced dimensionality space is horseshoe-like, and several trends are visible across the circuits. Some circuits show higher density across the outside edge of the horseshoe (e.g., L-R-RCPE-RCPE), and others show higher density on the inside edge of the horseshoe shape (RCPE-RCPE-RCPE-RCPE). Various circuits are more obviously separable by the side of the horseshoe, with RC-G-G, RC-RC-RCPE-RCPE, and Rs-Ws circuits having almost no samples on the higher right side of the horseshoe and dense clustering on the higher left and the L-R-RCPE-RCPE-RCPE circuit having almost no samples on the upper left side of the horseshoe. The L-R-RCPE circuit has a very high density at the upper right corner and very few spectra associated with the rest of the horseshoe. Other circuits show similar trends as those discussed, but overall, the clustering problem presented by this data set appears challenging and nonlinear. The overlap of component values after \gls{umap} dimensionality reduction across all circuit types indicates that the \gls{eis} measurements are not clearly differentiated by their \gls{ecm} label, and so identification of the correct \gls{ecm} label in this supervised learning problem may not be possible to achieve with high accuracy.

\subsection{Regression model}
\label{sec:si:reg}
An additional challenge of the hackathon was parameter estimation. While it was suggested that the finetuning of \gls{ecm} parameters should be done with the \textit{scipy} optimization framework \textit{curve\_fit}, the reliable suggestion of initial circuit parameters is essential. 

In the following, we present machine learning results based on a supervised approach. The results mentioned in this section are based on the original data set without filtering out any spectra as it was provided for the hackathon. We present it mainly to outline challenges and inspire others to develop new approaches. A key challenge is that the values of \gls{ecm} parameters used to generate the spectra differ by several orders of magnitude because they were mainly drawn from reciprocal distributions. This variation of parameter values across many orders of magnitude makes model training very difficult. The distribution of the values for all of the parameters for any given circuit model may be distributed non-normally. This intuition was validated by training regressors using the tsfresh-XGBoost pipeline and comparing it with a median dummy model. Here we use the same 80\%/20\% train/test split as used for the classification model. Separate models were trained for each circuit model, and multiple parameter values for each circuit model were predicted using the MultiOutputRegressor from \textit{sklearn}. Comparing this broad set of predictions, the dummy model often performed substantially better on the test data than the tsfresh-XGBoost pipeline.

Several methods for normalizing the distributions of the target variables were attempted: logarithmic transform, Yeo-Johnson power transform, and quantile transform using ten quantiles. Only the quantile transformation performed reliably better than the dummy model across the circuit types and parameters. Table \ref{tab:regression_tsfresh_xgboost_maes} reports mean absolute errors for regression models using the dummy model, the tsfresh-XGBoost model without any target variable transform, and the tsfresh-XGBoost model with quantile target transformation are reported for each circuit type and each parameter on the test split using. 

Furthermore, the results of the tsfresh-XGBoost models with and without quantile transformation change with changing random seed, suggesting that the regression task is brittle, i.e., minor changes in features and model can lead to considerable changes in prediction accuracy. The results in Tab.\ \ref{tab:regression_tsfresh_xgboost_maes} showcase thus only trends and difficulties with this task. Another question is how well the predictions perform as initial guesses in the subsequent \gls{ecm} parameter optimization. Further investigation and improvements or novel approaches are needed to tackle the abovementioned challenges.

{\small
\begin{longtable}[htb]{l|c|ccc}
\caption{Mean absolute error of parameter regression models on test splits for each ECM type, compared between dummy models that use the median values for each parameter in the training split, tsfresh-XGBoost regression models without any target data transformation, and tsfresh-XGBoost regression models using a quantile transformation on the target data. Bold entries correspond to the lowest value for each row.}
\label{tab:regression_tsfresh_xgboost_maes} \\

Circuit             & Parameter & Dummy             & tsfresh-XGBoost   & tsfresh-XGBoost-Quantile \\ \hline
\endfirsthead

Circuit             & Parameter & Dummy             & tsfresh-XGBoost   & tsfresh-XGBoost-Quantile \\ \hline
\endhead
           L-R-RCPE  &         L1  & 1.61E-06 & 1.81E-06 & \textbf{9.45E-07}\\
                     &         R1  & 4.58E+01 & \textbf{3.62E+01} & 3.69E+01\\
                     &         R2  & 3.92E+04 & 3.88E+04 & \textbf{2.70E+04}\\
                     &     CPE1\_t  & 1.17E-01 & 1.57E-02 & \textbf{1.48E-02}\\
                     &     CPE1\_C  & 4.67E-03 & 3.60E-03 & \textbf{3.59E-03}\\
      L-R-RCPE-RCPE  &         L1  & 1.49E-06 & 1.79E-06 & \textbf{1.11E-06}\\
                     &         R1  & \textbf{3.71E+01} & 4.54E+01 & 3.79E+01\\
                     &         R2  & 3.93E+02 & \textbf{3.22E+02} & 3.43E+02\\
                     &     CPE1\_t  & \textbf{1.24E-01} & 1.27E-01 & 1.29E-01\\
                     &     CPE1\_C  & \textbf{1.87E-06} & 3.57E-06 & 2.59E-06\\
                     &         R3  & 4.28E+04 & 5.36E+04 & \textbf{4.20E+04}\\
                     &     CPE2\_t  & 1.36E-01 & \textbf{2.88E-02} & 2.89E-02\\
                     &     CPE2\_C  & 1.47E-02 & 7.69E-03 & \textbf{6.96E-03}\\
 L-R-RCPE-RCPE-RCPE  &         L1  & 1.41E-06 & 1.73E-06 & \textbf{1.35E-06}\\
                     &         R1  & \textbf{3.69E+01} & 5.20E+01 & 4.02E+01\\
                     &         R2  & 1.90E+02 & 2.10E+02 & \textbf{1.73E+02}\\
                     &     CPE1\_t  & \textbf{1.32E-01} & 1.41E-01 & 1.40E-01\\
                     &     CPE1\_C  & \textbf{1.77E-06} & 3.21E-06 & 2.20E-06\\
                     &         R3  & 2.71E+03 & \textbf{1.63E+03} & 1.80E+03\\
                     &     CPE2\_t  & 1.36E-01 & \textbf{9.93E-02} & 1.00E-01\\
                     &     CPE2\_C  & 5.89E-04 & 7.05E-04 & \textbf{5.58E-04}\\
                     &         R4  & 9.87E+04 & 1.05E+05 & \textbf{9.38E+04}\\
                     &     CPE3\_t  & 1.26E-01 & \textbf{7.86E-02} & 8.38E-02\\
                     &     CPE3\_C  & 3.19E-02 & 3.01E-02 & \textbf{1.81E-02}\\
             RC-G-G  &         R1  & 5.18E+01 & \textbf{4.81E+01} & 5.27E+01\\
                     &         C1  & \textbf{5.38E-06} & 7.82E-06 & 7.38E-06\\
                     &       R\_g1  & \textbf{1.80E+00} & 4.71E+00 & 1.80E+00\\
                     &       t\_g1  & \textbf{6.12E-01} & 3.47E+00 & 6.13E-01\\
                     &       R\_g2  & 4.83E+01 & \textbf{4.77E+01} & 5.02E+01\\
                     &       t\_g2  & 1.67E+01 & 3.89E+01 & \textbf{1.57E+01}\\
    RC-RC-RCPE-RCPE  &         R1  & 2.88E+01 & 3.05E+01 & \textbf{3.33E+00}\\
                     &         R2  & 8.98E+00 & 1.05E+01 & \textbf{6.24E+00}\\
                     &         R3  & 3.50E+01 & 1.93E+01 & \textbf{4.82E+00}\\
                     &         R4  & \textbf{6.92E+04} & 2.68E+05 & 7.02E+04\\
                     &         C2  & 8.52E-06 & 1.27E-05 & \textbf{6.04E-06}\\
                     &     CPE3\_C  & 1.28E-04 & 3.58E-04 & \textbf{5.52E-05}\\
                     &     CPE4\_t  & 1.57E-01 & 1.77E-01 & \textbf{7.77E-02}\\
                     &     CPE4\_C  & 4.31E-01 & 5.57E-01 & \textbf{1.72E-01}\\
                     &         C1  & \textbf{0.00E+00} & 1.65E-16 & 5.66E-17\\
                     &     CPE3\_t  & \textbf{0.00E+00} & 8.34E-08 & 2.38E-08\\
          RCPE-RCPE  &         R1  & 4.24E+01 & \textbf{2.56E+01} & 3.29E+01\\
                     &         R2  & 3.71E+04 & 3.95E+04 & \textbf{2.34E+04}\\
                     &     CPE1\_t  & \textbf{1.20E-01} & 1.30E-01 & 1.27E-01\\
                     &     CPE1\_C  & \textbf{1.16E-03} & 1.36E-03 & 1.58E-03\\
                     &     CPE2\_t  & 1.26E-01 & \textbf{2.70E-02} & 2.91E-02\\
                     &     CPE2\_C  & 1.23E-02 & 9.52E-03 & \textbf{4.97E-03}\\
     RCPE-RCPE-RCPE  &         R1  & 1.19E+01 & 1.31E+01 & \textbf{1.16E+01}\\
                     &         R2  & 3.29E+02 & 5.24E+02 & \textbf{3.06E+02}\\
                     &         R3  & \textbf{7.34E+04} & 1.04E+05 & 7.42E+04\\
                     &     CPE1\_t  & \textbf{1.25E-01} & 1.55E-01 & 1.40E-01\\
                     &     CPE1\_C  & \textbf{1.74E-06} & 2.60E-06 & 1.74E-06\\
                     &     CPE2\_t  & 1.27E-01 & 1.26E-01 & \textbf{1.10E-01}\\
                     &     CPE2\_C  & 5.95E-05 & 1.47E-04 & \textbf{5.11E-05}\\
                     &     CPE3\_t  & 1.23E-01 & 1.15E-01 & \textbf{4.71E-02}\\
                     &     CPE3\_C  & 2.73E-02 & 2.10E-01 & \textbf{1.50E-02}\\
RCPE-RCPE-RCPE-RCPE  &         R1  & 3.93E+01 & 3.99E+01 & \textbf{3.74E+01}\\
                     &         R2  & 1.37E+02 & 1.84E+02 & \textbf{1.36E+02}\\
                     &         R3  & 2.50E+03 & 2.76E+03 & \textbf{2.29E+03}\\
                     &         R4  & 9.05E+04 & 1.52E+05 & \textbf{8.22E+04}\\
                     &     CPE1\_t  & \textbf{1.30E-01} & 1.36E-01 & 1.40E-01\\
                     &     CPE1\_C  & \textbf{4.55E-07} & 2.90E-04 & 4.56E-07\\
                     &     CPE2\_t  & 1.31E-01 & \textbf{1.26E-01} & 1.34E-01\\
                     &     CPE2\_C  & \textbf{8.80E-04} & 1.82E-03 & 8.82E-04\\
                     &     CPE3\_t  & 1.29E-01 & 1.06E-01 & \textbf{1.05E-01}\\
                     &     CPE3\_C  & \textbf{3.50E-03} & 6.71E-03 & 3.62E-03\\
                     &     CPE4\_t  & 1.30E-01 & \textbf{1.05E-01} & 1.06E-01\\
                     &     CPE4\_C  & 2.77E-02 & 3.06E-02 & \textbf{2.04E-02}\\
              Rs\_Ws  &         R1  & 1.00E+02 & \textbf{1.63E+01} & 2.63E+01\\
                     &       W1\_R  & 7.83E+02 & 2.59E+02 & \textbf{2.40E+02}\\
                     &       W1\_T  & \textbf{1.89E+02} & 1.94E+02 & 2.17E+02\\
                     &       W1\_p  & 9.69E-01 & 8.12E-02 & \textbf{8.08E-02}\\
\end{longtable}
} 

\subsection{CNN with colored images approach: MobileNetV2 architecture}
\label{sec:si:sup_info_cnn_freqColored}
During the BatteryDEV hackathon, one team  employed the MobileNetV2 architecture, available as part of the TensorFlow-Slim model library. MobileNetV2 is a neural network architecture with a lower computational cost at deployment relative to other CNNs with similar performance. It has been widely used for mobile and embedded vision applications. This neural network works based on a depthwise separable convolution, which has two steps. In the \textit{first step}, a depthwise convolution is applied to the input layer, whose output  acts as an intermediate set of values that will be the input of a \textit{second step}, which consists of a pointwise convolution \cite{https://doi.org/10.48550/arxiv.1704.04861}.

\begin{figure}[htb]
    \centering
    \includegraphics[width=.8\linewidth]{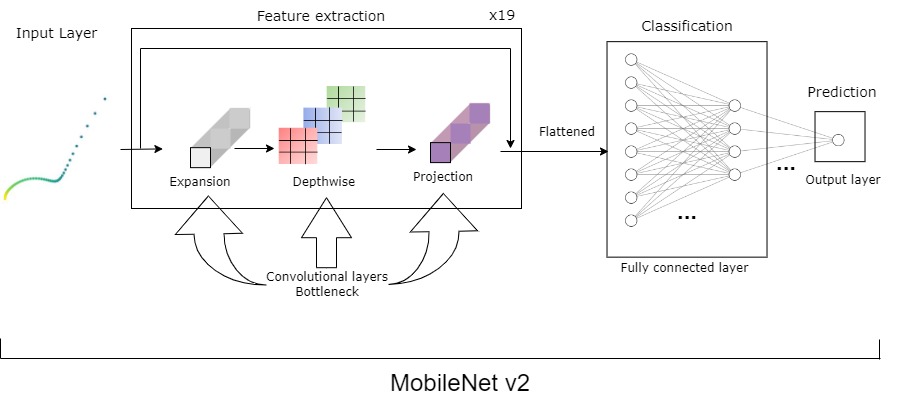}
    \caption{Architecture of MobileNetV2. MobileNetV2 works based on a depthwise separable convolution. The bottleneck block increases the size of the input layer's representation, allowing the neural network to learn more abstract and  complex relationships.}
    \label{fig:MobileNETV2}
\end{figure}

In this case, we used interpolated Nyquist plots obtained from the data set as the input layer for the MobileNetV2, as seen in the first layer of Fig.\ \ref{fig:MobileNETV2}.

The preprocessing step of this approach was similar to that described in Section \ref{sec:prepro_freq_range} for the XGBoost model. The spectra were interpolated to the shared frequency range (10$^1$ Hz -- 10$^5$ Hz). 
In addition, as the main reason to use a deep learning \gls{cnn} approach is to extract features from images, we constructed Nyquist plot figures (with real impedance at the x-axis and imaginary impedance at the y-axis) with an added color feature that could represent the associated frequency (ranging from yellow tones for high-frequencies to violet tones for low-frequencies) as can be seen in the example Nyquist plots in Fig.\ \ref{fig:MobileNETV2}. Finally, all the images were created in the same size of 128$\times$128 pixels.

Some potential reasons for this CNN approach's low and fluctuating accuracies compared to the other methods explored in this article could be the use of a scattered Nyquist plot instead of solid lines because it could make it harder for the CNN to extract features of the images. In addition, another challenge with colored images is to have all the RGB information in a unique image; some authors have proposed an alternative approach to enhance the performance of \gls{cnn} approaches. This alternate method involves achieving augmentations in the color space of the channels. Separating a channel of a certain color -- such as Red, Green, or Blue -- is a simple way to encode information in color \cite{alzubaidi2021review}. Furthermore, the architecture of \glspl{cnn}, can be sensitive to the thickness of the line or dots that represent the \gls{eis} spectra. Last the embeddings/features that we hope to extract from the \glspl{cnn} are very supposedly very different from those of the original image classification task of the MobileNETV2. Last, the amount of data that was available during this competition might not be sufficient to learn these features reliably.

\subsection{Bayesian Approach}
\label{sec:si:bayes}
The following Bayesian approach was not implemented during the hackathon and is presented here as an idea for future work. Future development in this direction is also desired as it can answer the regression and classification tasks in an unsupervised manner. Other metrics can be adopted to overcome this, such as Occam's razor \cite{rasmussen2000occam}, which can be computed as marginal likelihood with the Bayesian paradigm.

\paragraph{Bayesian inference}
In the context of Bayesian inference, the objective can be formulated as a parameter estimation task with
\begin{align}
    y_\text{noise} \sim \mathcal{N}(y; 0, \sigma_n), \\
    \textbf{y} = \text{ECM}(\textbf{X}) + y_\text{noise}, \\
    \ell(x) = \mathcal{N}(y_*; \textbf{y}, \sigma_n\textbf{I}), \\
    p(x|\textbf{y}) = \frac{\ell(x)\pi(x)}{Z}, \label{eq:bayes-a}\\
    Z = \int \ell(x)\pi(x) \text{d}x, \label{eq:bayes-b}
\end{align}
where $\textbf{X}$ is the parameter of the ground truth, $\textbf{y}$ is the observed \gls{eis}, $\sigma_n$ is the experimental noise variance, $\textbf{I}$ is the identity matrix, $\ell(x)$ is the likelihood, $\pi(x)$ is the prior, $p(x|\textbf{y})$ is the posterior, and $Z$ is the model evidence.
Therefore, the estimation of the parameter is the same as inferring posterior in Eq.\ \eqref{eq:bayes-a}. The classification criterion is given by the model evidence in Eq.\  \eqref{eq:bayes-b}. Such inferences can be simultaneously made with nested sampling \cite{skilling2006nested} or Bayesian quadrature \cite{adachi2022fast, adachi2022bayesian, adachi2023sober}. However, inferring the evidence from 10,000 models is computationally expensive. Bayesian quadrature can help to perform such inference tasks. 

\printbibliography[heading=subbibliography]
\end{refsection}

\end{document}